\PassOptionsToPackage{unicode}{hyperref}
\PassOptionsToPackage{hyphens}{url}

\documentclass[]{article}

\usepackage{iftex}
\usepackage{amsmath,amssymb}
\usepackage{bm}

\usepackage{float}
\usepackage{algorithm}
\usepackage{algpseudocode}

\algrenewcommand\algorithmicrequire{\textbf{Input:}}
\algrenewcommand\algorithmicensure{\textbf{Output:}}
\algrenewcommand\algorithmiccomment[1]{\hfill{\scriptsize$\triangleright$ #1}}

\usepackage{tikz}
\usetikzlibrary{
  arrows.meta,
  positioning,
  calc,
  shapes.geometric,
  fit,
  decorations.pathreplacing
}

\tikzset{
  sbox/.style={
    draw, rounded corners=2pt, align=center,
    minimum height=7mm, inner sep=3pt, line width=0.6pt
  },
  gbox/.style={
    draw, rounded corners=2pt, align=center,
    minimum height=7mm, inner sep=3pt, line width=0.6pt
  },
  arr/.style={-Stealth, line width=0.8pt}
}

\ifPDFTeX
  \usepackage[T1]{fontenc}
  \usepackage[utf8]{inputenc}
  \usepackage{textcomp}
  \usepackage{newtxtext,newtxmath}
  \usepackage{inconsolata}
\else
\fi

\IfFileExists{upquote.sty}{\usepackage{upquote}}{}
\IfFileExists{microtype.sty}{
  \usepackage[]{microtype}
  \UseMicrotypeSet[protrusion]{basicmath} 
}{}
\makeatletter
\@ifundefined{KOMAClassName}{
  \IfFileExists{parskip.sty}{%
    \usepackage{parskip}
  }{
    \setlength{\parindent}{0pt}
    \setlength{\parskip}{6pt plus 2pt minus 1pt}}
}{
  \KOMAoptions{parskip=half}}
\makeatother
\usepackage{xcolor}
\IfFileExists{xurl.sty}{\usepackage{xurl}}{} 
\IfFileExists{bookmark.sty}{\usepackage{bookmark}}{\usepackage{hyperref}}
\hypersetup{
  hidelinks,
  pdfcreator={LaTeX via pandoc}}
\usepackage[margin=1in]{geometry}
\usepackage{longtable,booktabs,array}
\usepackage{tabularx}   

\usepackage{calc} 
\usepackage{etoolbox}
\makeatletter
\patchcmd\longtable{\par}{\if@noskipsec\mbox{}\fi\par}{}{}
\makeatother
\IfFileExists{footnotehyper.sty}{\usepackage{footnotehyper}}{\usepackage{footnote}}
\makesavenoteenv{longtable}
\usepackage{graphicx}
\makeatletter
\def\maxwidth{\ifdim\Gin@nat@width>\linewidth\linewidth\else\Gin@nat@width\fi}
\def\maxheight{\ifdim\Gin@nat@height>\textheight\textheight\else\Gin@nat@height\fi}
\makeatother
\setkeys{Gin}{width=\maxwidth,height=\maxheight,keepaspectratio}
\makeatletter
\def\fps@figure{htbp}
\makeatother
\ifLuaTeX
  \usepackage{selnolig}  
\fi

\author{}
\date{}

\begin{document}
\begin{center}
DynamicGate‑MLP: Conditional Computation via Learned Structural Dropout and Input‑Dependent Gating for Functional Plasticity

Yong Il Choi

Sorynorydotcom Co., Ltd./AI Open Research Lab: \href{mailto:hurstchoi@sorynory.com}{\nolinkurl{hurstchoi@sorynory.com}}

\url{https://orcid.org/0009-0009-8813-5420}

December 24, 2025
\end{center}
\begin{abstract}
\noindent
Dropout is a representative regularization technique that stochastically deactivates hidden units during training to mitigate overfitting. In contrast, standard inference executes the full network with dense computation, so its goal and mechanism differ from conditional computation, where the executed operations depend on the input. This paper organizes DynamicGate-MLP into a single framework that simultaneously satisfies both the regularization view and the conditional-computation view. Instead of a random mask, the proposed model learns gates that decide whether to use each unit (or block), suppressing unnecessary computation while implementing sample-dependent execution that concentrates computation on the parts needed for each input. To this end, we define continuous gate probabilities and, at inference time, generate a discrete execution mask from them to select an execution path. Training controls the compute budget via a penalty on expected gate usage and uses a Straight-Through Estimator (STE) to optimize the discrete mask. We evaluate DynamicGate-MLP on MNIST, CIFAR-10, Tiny-ImageNet, Speech Commands, and PBMC3k, and compare it with various MLP baselines and MoE-style variants. Compute efficiency is compared under a consistent criterion using gate activation ratios and a layer-weighted relative MAC metric, rather than wall-clock latency that depends on hardware and backend kernels.
\end{abstract}
\section{Introduction}

\subsection{\textbf{1.1 Motivation}}

The starting point of this study dates back to the mid-2000s, when we encountered early neuromorphic chips such as General Vision and IBM's ZISC and began to suspect that ``brain-inspired'' chips might point beyond the traditional von Neumann architecture. Later, while training and analyzing large neural networks—especially LLMs—we repeatedly observed that dense computation in hidden layers is structurally far from biological mechanisms such as neuronal firing/silence and synaptic plasticity. We also found that sparse-activation phenomena such as firing/silence can be approximated more precisely in a mathematical form. Considering the current technical limitations of neuromorphic hardware, we believe there is a need for conditional-computation structures that are implementable on general-purpose hardware. In this context, we designed DynamicGate-MLP, which selectively performs computation via input-dependent gating.

\subsection{\textbf{1.2 Problem Statement}}

Deep learning models are often over-parameterized, which can be beneficial for expressivity and optimization but increases compute cost and may raise the risk of overfitting. Dropout is a widely used regularization method that removes random units during training to reduce co-adaptation and improve generalization{[}1{]}. However, standard dropout has the following limitations.

\begin{itemize}
\item Sparsity only during training \textbf{(Training-time sparsity only)}: inference is typically executed with dense computation, making it difficult to translate into conditional execution.
\item Input-agnostic stochasticity \textbf{(Input-agnostic stochasticity)}: in standard dropout, the mask is not chosen adaptively per input, but is sampled from fixed-probability randomness.

\end{itemize}

Pruning, on the other hand, can compress a model by removing weights/channels after training{[}14{]}, but it typically applies the same static structure to all inputs. Conditional computation and sparse routing aim to reduce average computation by executing only a subset of paths per input; Mixture-of-Experts (MoE){[}12{]} and Switch Transformer{[}13{]} are representative examples.

\subsection{\textbf{1.3 Core Idea}}

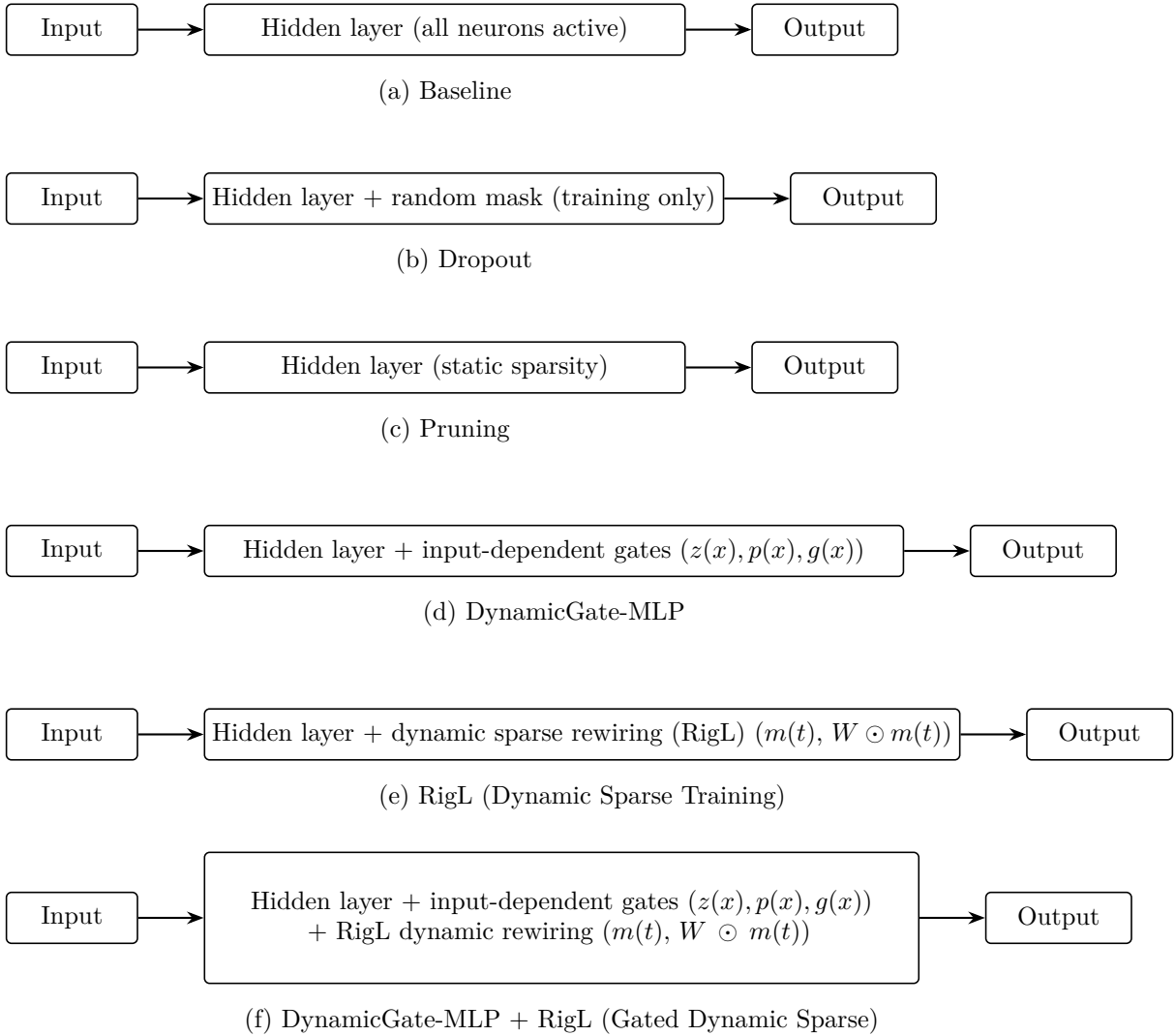
\begin{figure}[!htbp]
\centering
\begin{tikzpicture}[node distance=10mm]
  \node[sbox, minimum width=18mm] (inA) {Input};
  \node[gbox, right=9mm of inA, minimum width=66mm] (hidA) {Hidden layer (all neurons active)};
  \node[sbox, right=9mm of hidA, minimum width=20mm] (outA) {Output};
  \draw[arr] (inA) -- (hidA);
  \draw[arr] (hidA) -- (outA);
  \node[below=2mm of hidA] {(a) Baseline};

  \node[sbox, below=16mm of inA, minimum width=18mm] (inB) {Input};
  \node[gbox, right=9mm of inB, minimum width=66mm] (hidB) {Hidden layer + random mask (training only)};
  \node[sbox, right=9mm of hidB, minimum width=20mm] (outB) {Output};
  \draw[arr] (inB) -- (hidB);
  \draw[arr] (hidB) -- (outB);
  \node[below=2mm of hidB] {(b) Dropout};

  \node[sbox, below=16mm of inB, minimum width=18mm] (inC) {Input};
  \node[gbox, right=9mm of inC, minimum width=66mm] (hidC) {Hidden layer (static sparsity)};
  \node[sbox, right=9mm of hidC, minimum width=20mm] (outC) {Output};
  \draw[arr] (inC) -- (hidC);
  \draw[arr] (hidC) -- (outC);
  \node[below=2mm of hidC] {(c) Pruning};

  \node[sbox, below=18mm of inC, minimum width=18mm] (inD) {Input};
  \node[gbox, right=9mm of inD, minimum width=96mm] (hidD) {Hidden layer + input-dependent gates $(z(x),p(x),g(x))$};
  \node[sbox, right=9mm of hidD, minimum width=20mm] (outD) {Output};
  \draw[arr] (inD) -- (hidD);
  \draw[arr] (hidD) -- (outD);
  \node[below=2mm of hidD] {(d) DynamicGate-MLP};

\node[sbox, below=18mm of inD, minimum width=18mm] (inE) {Input};
\node[gbox, right=9mm of inE, minimum width=96mm] (hidE) {Hidden layer + dynamic sparse rewiring (RigL) $(m(t),\, W \odot m(t))$};
\node[sbox, right=9mm of hidE, minimum width=20mm] (outE) {Output};
\draw[arr] (inE) -- (hidE);
\draw[arr] (hidE) -- (outE);
\node[below=2mm of hidE] {(e) RigL (Dynamic Sparse Training)};


\node[sbox, below=18mm of inE, minimum width=18mm] (inF) {Input};

\node[gbox, right=9mm of inF,
      minimum width=96mm,
      text width=96mm,
      align=center,
      minimum height=18mm] (hidF)
{Hidden layer + input-dependent gates $(z(x),p(x),g(x))$\\
+ RigL dynamic rewiring $(m(t),\, W \odot m(t))$};

\node[sbox, right=9mm of hidF, minimum width=20mm] (outF) {Output};

\draw[arr] (inF) -- (hidF);
\draw[arr] (hidF) -- (outF);

\node[below=2mm of hidF] {(f) DynamicGate-MLP + RigL (Gated Dynamic Sparse)};

\end{tikzpicture}
\caption{Conceptual comparison: Baseline / Dropout / Pruning / DynamicGate-MLP / RigL / DynamicGate‑MLP + RigL}
\label{fig:conceptual}
\end{figure}
DynamicGate-MLP can control \emph{sample-wise activation patterns} via input-dependent gating. In Fig.~1, (d) and (f) are models newly designed in this work. This paper connects dropout--pruning--conditional computation into a single narrative through DynamicGate-MLP.

\begin{figure}[t]
\centering
\includegraphics[width=6.26806in,height=4.66111in]{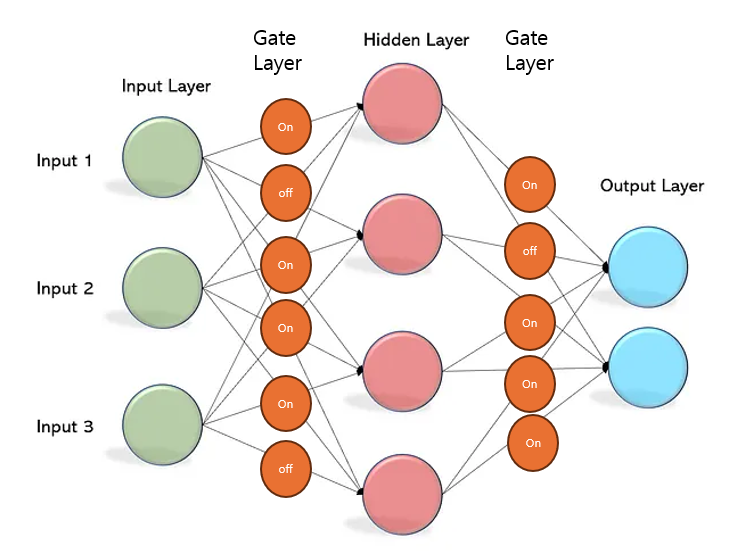}
\caption{Conceptual diagram of the gating layer}
\label{fig:concept_gatelayer}
\end{figure}

The core idea of DynamicGate-MLP is to implement the viewpoint: ``turn units off like dropout, but not randomly—turn them off by learning, and turn them off differently for each input at execution time.'' Concretely, we insert learnable gates into each layer, replacing dropout's random mask with a learned gate probability $p$. We then discretize this probability with a threshold to obtain a hard mask $g(x)$, allowing only selected units (or blocks) to participate in computation. At the same time, by including a penalty on gate usage (e.g., $\mathbb{E}[p]$) in the objective, we can directly tune the activation-rate (compute) budget during training while maintaining accuracy.
Furthermore, combining grow-and-prune methods such as RigL/SET enables an extension that couples fast time-scale functional selection (``which units to use for this input'') with slow time-scale structural change (``which connections should exist at all''). From a neuroscience perspective, the brain exhibits both functional plasticity—selectively activating circuits depending on tasks and context—and structural plasticity—reconfiguring circuits via synapse/spine formation and elimination{[}27,28{]}. DynamicGate-MLP focuses primarily on the former (functional selection), while its combination with RigL-style rewiring can complementarily connect it to the latter (structural change).

\subsection{\textbf{1.4 Contributions}}

\begin{itemize}

\item We introduce a shared gating structure that decides unit activation, unifying dropout-style probabilistic masking (during training) and input-dependent conditional execution (at inference) within a single gating layer.

\item We introduce expected gate-usage regularization to directly control a compute budget and provide an implementable training method that stably learns a discrete gate policy via STE.

\item We compare Baseline / Dropout / Pruned / DynamicGate-MLP / RigL / DynamicGate-MLP + RigL on MNIST and CIFAR.

\item We report compute using proxy metrics based on gate activation ratios (Compute Proxy) and a MAC-weighted relative metric (RelMAC), and we do not equate these directly with wall-clock latency, which depends on backend optimizations.
\end{itemize}

\section{\textbf{2. Background and Related Work}}

\subsection{\textbf{2.1 Neuronal firing, silence, and removal as conditional computation}}

\begin{figure}[t]
\centering
\includegraphics[width=6.26806in,height=4.66111in]{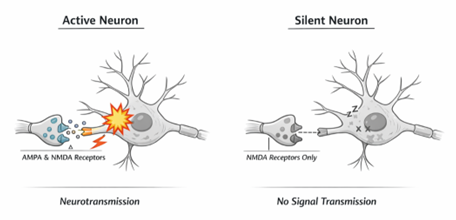}
\caption{Conceptual illustration of active and silent neurons}
\label{fig:silent_neuron}
\end{figure}

\begin{quote}
In neuroscience, neuronal silence refers to a state where a neuron structurally exists but does not fire under certain conditions. This can be understood as part of sparse information processing in which we do not ``always compute everything'' but rather ``compute only when needed.'' Over longer time scales, it is also related to structural changes such as pruning, where rarely used connections weaken or are removed. This view can be summarized as follows.
\end{quote}

\begin{itemize}
\item Short term: input-dependent firing/silence (functional selection)

\item Long term: removal/rewiring based on usage statistics (structural change)
\end{itemize}

\begin{quote}
DynamicGate-MLP implements such ``reversible silence'' via input-dependent gating, (reversible silence)\footnote{Here, ``reversible silence'' means that a unit/connection is not permanently removed, but is temporarily deactivated depending on the input and thus excluded from computation. Therefore, an inactive unit can become active again for future inputs, and once re-activated, the parameters on that path re-enter gradient updates from that point onward.} and it provides a bridge between gating and pruning in that units with low average gate usage can become candidates for long-term removal.
\end{quote}

\subsection{\textbf{2.2 Dropout, DropConnect, and Bayesian interpretations}}

Dropout{[}1{]} and DropConnect{[}2{]} stochastically deactivate units or connections during training to improve generalization\footnote{I.e., improving performance on unseen inputs.}. Dropout can also be interpreted from a Bayesian approximation perspective{[}4{]}, and many extensions exist, including variational dropout{[}3{]} and Concrete Dropout{[}5{]}. DynamicGate-MLP can be viewed as a learnable/structured variant of dropout in that it uses 	extbf{learned gate probabilities} instead of a fixed-probability random mask.

\subsection{\textbf{2.3 Learned sparsity, discrete gating, and continuous relaxations}}

L0-regularization-based sparse learning{[}7{]} learns structural sparsity by directly controlling the expected activation of a mask. To make discrete choices differentiable, various methods have been proposed, including STE{[}6{]}, Gumbel-Softmax{[}8{]}, and Concrete distributions{[}9{]}. For simplicity and clarity of implementation, this paper focuses on hard threshold-based discrete gates trained with STE.

\subsection{\textbf{2.4 Pruning, structured sparsity, and hardware friendliness}}

Pruning compresses a model by removing weights/channels/filters{[}14{]}. The Lottery Ticket Hypothesis{[}15{]} and Movement Pruning{[}16{]} discuss the existence of learnable sparse structures and training strategies. Structured pruning (at the channel/filter level){[}17{]} is often favorable for real speedups, but it typically yields a static structure with weak input dependence.

\subsection{\textbf{2.5 Conditional computation, MoE, and adaptive computation}}

Conditional computation reduces average computation by executing only a subset of paths for each input. ACT{[}11{]}, MoE{[}12{]}, and Switch Transformer{[}13{]} are representative approaches. Unlike MoE, which introduces multiple experts, DynamicGate-MLP implements selective execution of units (or blocks) inside a simple MLP, aiming for a compact formulation that explicitly includes budget control.

\subsection{\textbf{2.6 Dynamic sparse training (DST) and rewiring}}

SET{[}20{]} and RigL{[}30{]} are dynamic sparse training (DST) methods that prune and regrow connections during training, offering a rewiring perspective in which the structure itself changes over time. While this paper primarily focuses on functional gating, we later present an extension that combines gating with rewiring.

\section{\textbf{3. DynamicGate‑MLP: Unified Formulation}}

Consider an $L$-layer MLP. For an input $x$, let $h^{(\ell-1)}(x)$ denote the activation vector of layer $\ell-1$. A standard linear layer is defined as follows.

\begin{equation}\label{eq:mlp_def}
\begin{aligned}
z^{(l)}(x) &= W^{(l)}h^{(l-1)}(x) + b^{(l)}, \\
h^{(l)}_{\mathrm{raw}}(x) &= \phi\!\bigl(z^{(l)}(x)\bigr)
\end{aligned}
\end{equation}

DynamicGate-MLP multiplies each hidden unit by a gate.

\begin{equation}\label{eq:mlp_def}
\begin{aligned}
h^{(l)}(x) &= g^{(l)}(x)\odot h^{(l)}_{\mathrm{raw}}(x).
\end{aligned}
\end{equation}
.

Here, $g^{(l)}(x)$ is defined as follows:
\begin{equation}\label{eq:mlp_def}
g^{(l)}(x)\in[0,1]\quad(\text{soft gate}),\qquad
g^{(l)}(x)\in\{0,1\}\quad(\text{hard gate}).
\end{equation}
That is, a soft gate is a real value between 0 and 1, while a hard gate takes either 0 or 1.

\subsection{\textbf{3.1 Static vs. dynamic gating}}

• Static (input-agnostic): for each unit, introduce a learnable logit $\mathcal{Z}_{i}^{(l)}$ and produce a fixed probability $\mathcal{P}_{i}^{(l)}$. This can be interpreted as learned structural dropout.

• Dynamic (input-dependent): using a small gate network.

\textbf{(1) Logit (score) generation}

\begin{equation}\label{eq:mlp_def}
z_g^{(l)}(x) = \mathrm{GateNet}\!\left(h^{(l-1)}(x)\right)
\end{equation}

\begin{itemize}
\item
  \(h^{\left( l-1 \right)}(x)\): representation from the previous layer (feature vector)
\item
  \(GateNet(\cdot)\): a small network that decides ``which neurons/blocks to open, and by how much, for this input.''
\end{itemize}

This naturally connects to conditional computation {[}11, 12{]}.

\(z_{g}^{(l)}(x)\): logit = the gate's ``raw score''

\begin{itemize}
\item
  Value range: \(( - \infty, + \infty)\)
\item
  Meaning: learned toward large positive values if it should open, and toward large negative values if it should close
\end{itemize}

\textbf{(2) Converting to gate probabilities with a sigmoid}

\begin{itemize}
\item
  \(\sigma(t) = \scalebox{1.2}{$\displaystyle \frac{1}{1 + e^{- t}}$}\) (sigmoid)
\item
  Result: \(g^{(l)}(x) \in (0,1)\)
\item
  Interpretation:

  \begin{itemize}
  \item
    \(z_{g} \gg 0 \Rightarrow g \approx 1\) (almost fully ON)
  \item
    \(z_{g} \ll 0 \Rightarrow g \approx 0\) (almost fully OFF)
  \item
    \(z_{g} = 0 \Rightarrow g = 0.5\)
  \end{itemize}
\end{itemize}

In other words, this step converts GateNet's score \(z_g\) into something like the probability/fraction that the gate is ON.

\subsection{\textbf{3.2 Probability parameterization, threshold, and temperature}}

From the gate logit \(\mathcal{Z}_{i}^{(l)}(x)\), we define

\begin{equation}\label{eq:mlp_def}
p_{i}^{(\ell)}(x)
= \sigma\!\left(\frac{z_{g,i}^{(\ell)}(x)}{\tau}\right),
\qquad
g_{i}^{(\ell)}(x)
= \mathbb{1}\footnote{$\mathbb{1}$: the indicator function.}\!\left[p_{i}^{(\ell)}(x) > \theta\right].
\end{equation}

where $\tau$ is the temperature and $\theta$ is a global hard threshold. (Extension: if layerwise thresholds are needed, this can be generalized to $\theta^{(l)}$, but for simplicity we use a global $\theta$.)

For stability, training may use soft gating ($p$), while the deployment/inference policy is often defined by hard gating ($g$). Accordingly, we distinguish the following when reporting results.

• Training proxy: $\mathbb{E}[\emph{p}]$ (differentiable expected activation)

• Deployment activation: $\mathbb{E}[g] = \mathbb{E}[1(\emph{p} > \theta)]$

\subsection{\textbf{3.3 Budget control via expected gate usage}}

We denote the total objective by $J$ to avoid confusion with the number of layers $L$.
\begin{equation}\label{eq:mlp_def}
J
= L_{\text{task}}
+ \lambda_g \sum_{\ell=1}^{L-1} \frac{1}{n_\ell}\sum_{i=1}^{n_\ell} \bar{p}_{i}^{(\ell)},
\qquad
\bar{p}_{i}^{(\ell)}
= \frac{1}{|B|}\sum_{x\in B} p_{i}^{(\ell)}(x).
\end{equation}

\textbf{(1) Expected activation ratio of layer \(l\) (definition)}

\begin{equation}\label{eq:mlp_def}
\rho^{(l)}\text{\:\,}: = \text{\:\,}\frac{1}{n_{l}}\sum_{i = 1}^{n_{l}}{\mathbb{E}\text{ }\text{⁣}}\left\lbrack p_{i}^{(l)}(x) \right\rbrack\text{\:\,} \approx \text{\:\,}\frac{1}{n_{l}}\sum_{i = 1}^{n_{l}}\frac{1}{\mid B \mid}\sum_{x \in B}^{}p_{i}^{(l)}(x)
\end{equation}

  \(\rho^{(l)}\) indicates ``how much (probabilistically) layer \(l\) is ON on average.''

\vspace{1\baselineskip}
\textbf{(2) The corresponding penalty term in the objective}

\begin{equation}\label{eq:mlp_def}
\lambda_{g}\sum_{l = 1}^{L - 1}\rho^{(l)}
\end{equation}

This is similar to L0-style sparse learning objectives {[}7{]}.

\textbf{(3) Compute-cost weighting (optional).} If we introduce a per-unit compute cost \(C_{i}^{(l)}\), then

\begin{equation}\label{eq:mlp_def}
J
= L_{\text{task}}
+ \lambda_g \sum_{\ell=1}^{L-1}
\frac{\sum_{i=1}^{n_\ell} c_i^{(\ell)}\, p_i^{(\ell)}}
{\sum_{i=1}^{n_\ell} c_i^{(\ell)}}.
\end{equation}

which enables cost-aligned budget control {[}17{]}.

\vspace{1\baselineskip}
\subsection{\textbf{3.4 Top‑k gating (optional hard budget)}}

Instead of a threshold, one can activate exactly \emph{$k_\ell$} units per layer:

\begin{equation}\label{eq:mlp_def}
g^{(\ell)}(x) = \operatorname{TopK}\!\left(p^{(\ell)}(x),\, k_\ell\right).
\end{equation}

enforces exactly \(k_{\ell}\) activations per layer to guarantee a strict budget, but it may reduce the flexibility of per-input adaptive control {[}13{]}.

\subsection{\textbf{3.5 Learning discrete gates via STE}}

Because hard gates are non-differentiable, we use STE {[}6{]}. The forward pass uses \emph{g}, while the backward pass uses the derivative of \emph{p}.

\textbf{STE equations (hard gate in forward, soft-gate gradient in backward)}

\textbf{(1) Soft gate probability (differentiable path)}

\begin{equation}\label{eq:mlp_def}p_{i}^{\left( \mathcal{l} \right)}(x) = \sigma\text{ }\text{⁣}\left( \frac{z_{g,i}^{\left( \mathcal{l} \right)}(x)}{\tau} \right)
\end{equation}

\textbf{(2) Forward: use a hard gate (binarization)}

\begin{equation}\label{eq:mlp_def}g_{i}^{\left( \mathcal{l} \right)}(x) = \mathbb{1}\text{ }\text{⁣}\left\lbrack p_{i}^{\left( \mathcal{l} \right)}(x) > \theta \right\rbrack
\end{equation}

\textbf{(3) Backward(STE):}
\(\frac{\partial \mathbf{g}}{\partial \mathbf{z}_{g}}\)\textbf{ is approximated by}
\(\frac{\partial \mathbf{p}}{\partial \mathbf{z}_{g}}\)\textbf{.}

\begin{equation}\label{eq:mlp_def}\frac{\partial g_{i}^{\left( \mathcal{l} \right)}(x)}{\partial z_{g,i}^{\left( \mathcal{l} \right)}(x)}\text{\:\,} \approx \text{\:\,}\frac{\partial p_{i}^{\left( \mathcal{l} \right)}(x)}{\partial z_{g,i}^{\left( \mathcal{l} \right)}(x)}\text{\:\,} = \text{\:\,}\frac{1}{\tau}\text{ }p_{i}^{\left( \mathcal{l} \right)}(x)(1 - p_{i}^{\left( \mathcal{l} \right)}(x))
\end{equation}

(If we write it in chain-rule form including the loss \(J\))

\begin{equation}\label{eq:mlp_def}\frac{\partial J}{\partial z_{g,i}^{\left( \mathcal{l} \right)}(x)}\text{\:\,} \approx \text{\:\,}\frac{\partial J}{\partial g_{i}^{\left( \mathcal{l} \right)}(x)} \cdot \frac{1}{\tau}\text{ }p_{i}^{\left( \mathcal{l} \right)}(x)(1 - p_{i}^{\left( \mathcal{l} \right)}(x))
\end{equation}

Alternatively, one may use Gumbel-Softmax {[}8{]} or Concrete {[}9{]}.

\begin{figure}[t]
\centering
\resizebox{\linewidth}{!}{%
\begin{tikzpicture}[
  node distance=7mm,
  gbox/.style={draw, rounded corners=3pt, thick, fill=gray!15, align=center, inner sep=5pt},
  arr/.style={-Latex, thick},
  every node/.style={font=\small}
]

\node[gbox, text width=22mm] (Ltask) {$\mathcal{L}_{\text{task}}$};

\node[gbox, text width=28mm, right=10mm of Ltask] (y)
{$y_i = g_i \cdot h_i$};

\node[gbox, text width=56mm, align=left, right=10mm of y] (ste)
{\textbf{Forward pass:}\\
$p_i(x)=\sigma\!\left(\frac{z_i(x)}{\tau}\right)$\\
$g_i(x)=\mathbf{1}\!\left[p_i(x)>\theta\right]$\\[1mm]
\textbf{Backward pass (STE):}\\
$\dfrac{\partial g_i}{\partial z_i}\approx
\dfrac{\partial p_i}{\partial z_i}=
\dfrac{1}{\tau}p_i(1-p_i)$};

\node[gbox, text width=18mm, right=10mm of ste] (z)
{$z_i(x)$};

\draw[arr] (Ltask) -- node[midway, above] {$\frac{\partial \mathcal{L}}{\partial y_i}$} (y);
\draw[arr] (y) -- node[midway, above] {$\frac{\partial \mathcal{L}}{\partial g_i}$} (ste);

\draw[arr] (ste) -- node[midway, above, yshift=2mm, align=center] {Surrogate\\gradient} (z);

\node[below=12mm of ste] (pen) {$\lambda_g \sum_i \bar{p}_i$};
\draw[arr] (pen) -- (ste.south);

\draw[arr]
  (z.south) -- ++(0,-20mm)
  -| ($(y.south)+(0,-10mm)$)
  -- (y.south);

\end{tikzpicture}%
}
\caption{Conceptual diagram of forward execution with hard gating and STE-based backpropagation}
\label{fig:ste}
\end{figure}
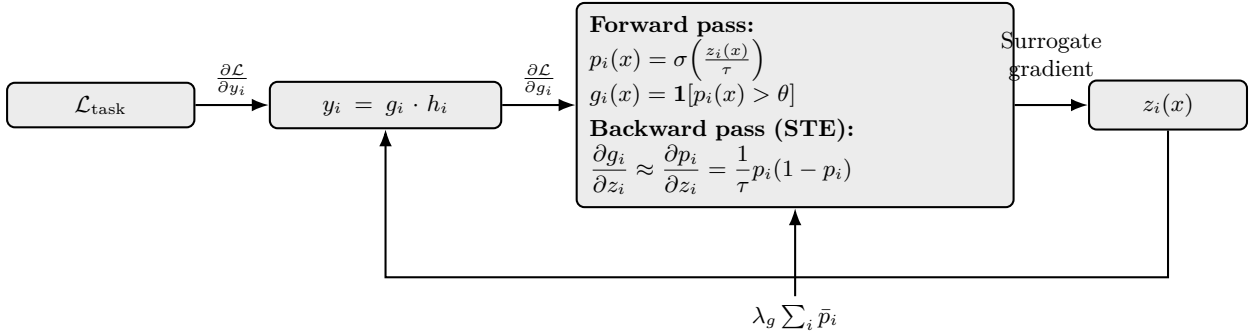

A gate-usage penalty enables budget control. Fig.~4 illustrates the concept of learning discrete gates with STE. Budget control is performed via a penalty based on $\mathbb{E}[\emph{p}]$, while deployment-time activation is interpreted via $\mathbb{E}[\emph{g}]$.

\subsection{\textbf{3.6 The gap between compute proxies and real deployment speed}}

In this paper, compute reduction is not reported via direct measurements of FLOPs/latency, but via proxy metrics based on gate activations. A simple average unit-activation proxy is defined as follows.

\begin{equation}\label{eq:mlp_def}
\alpha_p^{(\ell)}=\frac{1}{n_\ell}\sum_{i=1}^{n_\ell} p_i^{(\ell)},
\qquad
\alpha_g^{(\ell)}=\frac{1}{n_\ell}\sum_{i=1}^{n_\ell}\mathbb{E}\!\left[\mathbf{1}\!\left(p_i^{(\ell)}>\theta\right)\right].
\end{equation}

\begin{equation}\label{eq:mlp_def}
\mathrm{ComputeProxy}_p
= \frac{1}{L-1}\sum_{\ell=1}^{L-1}\alpha_p^{(\ell)},
\qquad
\mathrm{ComputeProxy}_g
= \frac{1}{L-1}\sum_{\ell=1}^{L-1}\alpha_g^{(\ell)}.
\end{equation}

However, MACs of fully connected (FC) layers are weighted by fan-in/out, so

we additionally define a relative MAC (RelMAC) metric:

\begin{equation}\label{eq:mlp_def}
\mathrm{RelMAC}
=
\frac{\sum_{\ell=1}^{L-1}\alpha^{(\ell)}\, n_{\ell-1}n_\ell}
{\sum_{\ell=1}^{L-1} n_{\ell-1}n_\ell},
\qquad
\alpha^{(\ell)} \in \{\alpha_p^{(\ell)},\, \alpha_g^{(\ell)}\}.
\end{equation}

DynamicGate controls input-dependent unit activation $\emph{g}(\emph{x})$, while RigL periodically rewires a sparse connectivity mask $\emph{m}(\emph{t})$ during training to learn structural sparsity.

Here, $\alpha^{(\ell)}$ denotes the layerwise average activation (under either the training proxy or the deployment policy). Real speedups depend on sparse kernels, block/channel-wise structured sparsity, or routing-based implementations {[}17, 14{]}. This paper focuses on controllable activation patterns and accuracy rather than absolute speed.

\subsection{\textbf{3.7 RigL: learning dynamic sparse connectivity (rewiring)}}

RigL is a Dynamic Sparse Training method that dynamically reallocates the connections themselves (a sparse mask over weights) during training, maximizing performance under a fixed parameter budget{[}30{]}.

\textbf{(1) Sparse-mask parameterization.}

For the layer-$\ell$ weight matrix $W^{(\ell)} \in \mathbb{R}^{n_\ell \times n_{\ell-1}}$, define a discrete mask at training step $t$

\begin{equation}\label{eq:mlp_def}
m^{(\ell)}(t) \in \{0,1\}^{n_\ell \times n_{\ell-1}}.
\end{equation}

and define the sparse weights used in computation as
\begin{equation}\label{eq:mlp_def}
\widetilde{W}^{(\ell)}(t) = W^{(\ell)}(t) \odot m^{(\ell)}(t).
\end{equation}

as follows. Here, ${\widetilde{W}}^{\left( \mathcal{l} \right)}(t)$ is the ``mask-applied weight actually used in computation.'' Note that the mask $m(t)$ is not updated at every step; under RigL it is updated every fixed period $(\Delta T)$, and remains fixed between updates while only $W(t)$ is trained. Then the forward computation of layer $\ell$ is

\begin{equation}\label{eq:mlp_def}
a^{(\ell)}(x;t)=\widetilde{W}^{(\ell)}(t)\,h^{(\ell-1)}(x;t)+b^{(\ell)},
\qquad
h^{(\ell)}(x;t)=\phi\!\big(a^{(\ell)}(x;t)\big).
\end{equation}

\textbf{(2) Fixed sparsity constraint.} To keep the number of active connections per layer constant, the mask is maintained to satisfy the following budget condition.

\begin{equation}\label{eq:mlp_def}
\left\|m^{(\ell)}(t)\right\|_{0} = (1 - s_{\ell})\, n_{\ell}\, n_{\ell-1}.
\end{equation}

Here, $s_{\ell}\in[0,1]$ is the layerwise sparsity; RigL keeps the number of connections fixed while changing which connections remain active during training.

\textbf{(3) Prune \& Grow (RigL update).} RigL updates the mask every period $\Delta T$.
\begin{center}
   t ∈ \{ΔT, 2ΔT, 3ΔT, . . . \}. (15)
\end{center}
At each update, it prunes (removes) active connections ($m=1$) with small weight magnitude and grows (adds) new connections among inactive locations ($m=0$) with large gradient magnitude.

In equations, define the set of indices to prune as

\begin{equation}\label{eq:mlp_def}
\mathcal{P}^{(\ell)}(t)
= \mathrm{BottomK}\!\Big(
\big\{\, |W_{ij}^{(\ell)}(t)| \;:\; m_{ij}^{(\ell)}(t)=1 \,\big\},
\, K_{\ell}
\Big).
\end{equation}

The set of indices to grow (new connections) is the Top-$K$ gradients among locations with no current connection (\(m=0\)).

\begin{equation}\label{eq:mlp_def}
\mathcal{G}^{(\mathcal{l})}(t)
= \operatorname{TopK}\!\Big(
\big\{\, |\nabla_{W_{ij}^{(\mathcal{l})}(t)} \mathcal{L}(t)| \;:\; m_{ij}^{(\mathcal{l})}(t)=0 \,\big\},
K_{\mathcal{l}}
\Big).
\end{equation}

\noindent\hspace*{3.5cm}%
\begin{minipage}{2.7\linewidth}
\raggedright
\begin{itemize}
\item
  \(\mathcal{L}(t)\) or \(L_{\text{task}}(t)\): the loss
\item
  \( \nabla_{W_{ij}}\mathcal{L}\): how sensitive the loss would be to that connection if it existed
\item
  \(m_{ij}(t)=0\): crucially, we only select candidates among currently absent connections
\end{itemize}
\end{minipage}
Define them as above, and update the mask as

\begin{equation}\label{eq:mlp_def}
m_{ij}^{(\ell)}(t^{+}) =
\begin{cases}
0, & (i,j)\in \mathcal{P}^{(\ell)}(t),\\
1, & (i,j)\in \mathcal{G}^{(\ell)}(t),\\
m_{ij}^{(\ell)}(t), & \text{otherwise}.
\end{cases}
\end{equation}

as follows. Here, $t^{+}$ denotes immediately after the update. This procedure preserves the parameter budget while allowing training to discover important connections {[}30{]}.

\textbf{(4) RigL compute proxy.} The layerwise connection density in RigL is

\begin{equation}\label{eq:mlp_def}
p_{\ell} = \frac{\left\lVert m^{(\ell)}(t)\right\rVert_{0}}{n_{\ell}\,n_{\ell-1}}
= 1 - s_{\ell}.
\end{equation}

and can be used as a proxy for structural (connection-level) sparsity.

\subsection{\textbf{3.8 DynamicGate‑MLP + RigL: a unified formulation of gating and rewiring}}

\begin{figure}[H]
\centering
\resizebox{\linewidth}{!}{%
\begin{tikzpicture}[
  node distance=12mm and 14mm,
  gbox/.style={draw, rounded corners=3pt, thick, fill=gray!15,
               align=center, inner sep=7pt, minimum height=10mm},
  sbox/.style={draw, rounded corners=3pt, thick, fill=gray!10,
               align=center, inner sep=6pt, minimum height=9mm},
  arr/.style={-Latex, thick},
  every node/.style={font=\small}
]
\node[sbox, text width=20mm] (Ltask) {$\mathcal{L}_{\text{task}}$};
\node[sbox, text width=30mm, minimum height=18mm, inner sep=8pt, right=18mm of Ltask] (y)
{\raisebox{8mm}{$y_i = g_i\cdot h_i$}};

\node[gbox, text width=66mm, align=left, right=18mm of y] (ste)
{\textbf{Gating forward pass:}\\
$p_i(x)=\sigma\!\left(\dfrac{z_i(x)}{\tau}\right)$\\
$g_i(x)=\mathbf{1}\!\left[p_i(x)>\theta\right]$\\[1mm]
\textbf{STE backward pass:}\\
$\dfrac{\partial g_i}{\partial z_i}\approx
\dfrac{\partial p_i}{\partial z_i}=
\dfrac{1}{\tau}p_i(1-p_i)$};

\node[sbox, text width=18mm, right=18mm of ste] (z) {$z_i(x)$};

\draw[arr] (Ltask) -- node[midway, above] {$\frac{\partial \mathcal{L}}{\partial y_i}$} (y);
\draw[arr] (y) -- node[midway, above] {$\frac{\partial \mathcal{L}}{\partial g_i}$} (ste);
\draw[arr] (ste) -- node[midway, above, yshift=2mm, align=center] {Surrogate\\gradient} (z);

\node[gbox, text width=30mm, below=44mm of ste] (pen) {$\lambda_g \sum_i \bar{p}_i$};
\draw[arr] (pen.north) -- (ste.south);

\node[sbox, text width=18mm, below=28mm of Ltask] (W) {$W$};
\node[sbox, text width=18mm, below=12mm of W] (m) {$m(t)$};

\node[gbox, text width=50mm, right=18mm of W] (Wtilde)
{$\widetilde{W}(t)=W\odot m(t)$\\
\footnotesize(sparse weights)};

\node[gbox, text width=55mm, align=left, right=22mm of Wtilde] (rigl)
{\textbf{RigL update (periodic):}\\
Prune: Bottom-$K$ by $|W|$\\
Grow: Top-$K$ by $|\nabla_W \mathcal{L}|$\\
$\|m(t)\|_0$ kept fixed};

\draw[arr] (W.east) -- (Wtilde.west);
\draw[arr] (m.east) -- (Wtilde.west);

\draw[arr] (rigl.west) to[bend left=35] node[midway, below] {rewire} (m.east);

\coordinate (gmid) at ($(Ltask.south)+(0,-18mm)$);
\draw[arr] (Ltask.south) -- (gmid) -| node[pos=0.35, above] {$\nabla_W\mathcal{L}$} (Wtilde.north);

\coordinate (upW) at ($(Wtilde.north)+(0,16mm)$);
\draw[arr] (Wtilde.north) -- (upW) -| node[pos=0.8, above, yshift=4mm] {$h_{\text{raw}}=\phi(\widetilde{W}x)$} (y.south);


\coordinate (zdown) at ($(z.south)+(0,-15mm)$);
\coordinate (ydown) at ($(y.south)+(0,-11mm)$);
\draw[arr] (z.south) -- (zdown) -| (ydown) -- (y.south);

\end{tikzpicture}%
}
\caption{Diagram of adding RigL's dynamic rewiring to the hard-gating (STE) training path. DynamicGate controls input-dependent unit activation $g(x)$, and RigL periodically rewires a sparse connectivity mask $m(t)$ during training to learn structural sparsity.}
\label{fig:ste_rigl_added_clean}
\end{figure}
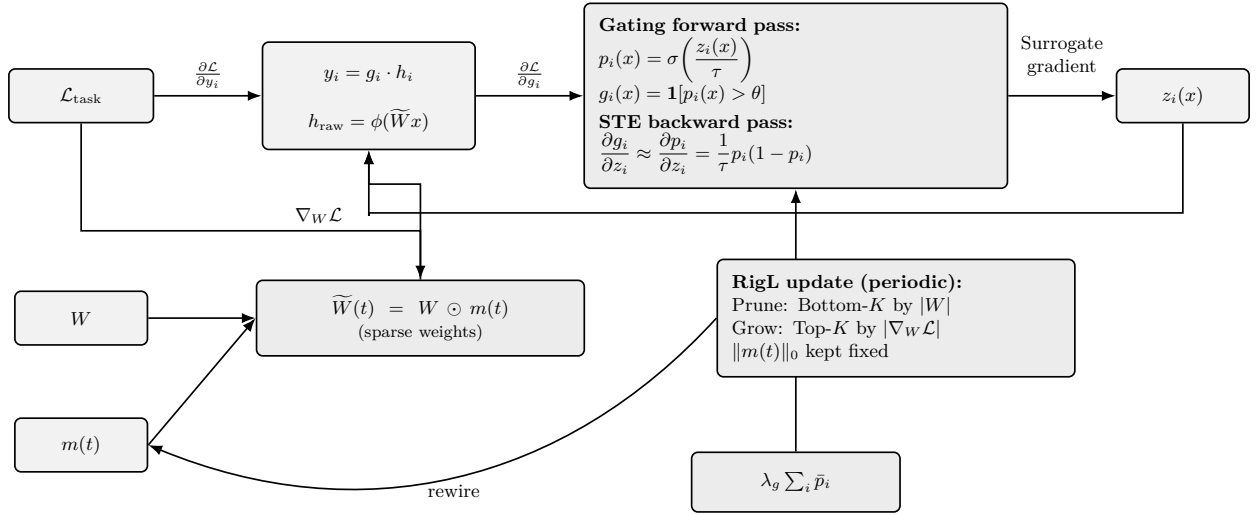

In this subsection, we define a fused model that applies input-dependent unit gating (DynamicGate-MLP) and connection rewiring (RigL) simultaneously. The key point is that (1) unit-level conditional computation and (2) connection-level structural learning provide sparsity along different axes.

\textbf{(1) Fused forward computation.} 

With RigL, the sparse weight is $\widetilde{W}^{(\ell)}(t)=W^{(\ell)}(t)\odot m^{(\ell)}(t)$, and

\textbf{(Sparse weight definition)}

\begin{equation}\label{eq:mlp_def}{\widetilde{W}}^{\left( \mathcal{l} \right)}(t) = W^{\left( \mathcal{l} \right)}(t) \odot m^{\left( \mathcal{l} \right)}(t)
\end{equation}

\textbf{(Forward: sparse structure + unit gating)}
\begin{equation}\label{eq:mlp_def}
a^{(\ell)}(x;t)=\tilde{W}^{(\ell)}(t)\,h^{(\ell-1)}(x;t)+b^{(\ell)}.
\end{equation}

\begin{equation}\label{eq:mlp_def}
h_{\mathrm{raw}}^{(\ell)}(x;t)=\phi\!\left(a^{(\ell)}(x;t)\right).
\end{equation}

\begin{equation}\label{eq:mlp_def}
h^{(\ell)}(x;t)=g^{(\ell)}(x;t)\odot h_{\mathrm{raw}}^{(\ell)}(x;t).
\end{equation}

Here, the mask \(m^{(\ell)}(t)\) constrains the \emph{connections (structure)}, and
the gate \(g^{(\ell)}(x;t)\) selects \emph{unit activation (function)} depending on the input.
\par\vspace{1\baselineskip}

\begin{equation}\label{eq:mlp_def}
h^{(\ell)}(x;t)
= g^{(\ell)}(x;t)\odot
\phi\!\left(
\bigl(W^{(\ell)}(t)\odot m^{(\ell)}(t)\bigr)\,h^{(\ell-1)}(x;t)
+ b^{(\ell)}
\right).
\end{equation}
\par\vspace{1\baselineskip}

In short, connectivity is constrained by the mask, and unit activation is selected per input on top of it.

GateNet input: the gate logits are defined in the same way as before,

\begin{equation}\label{eq:mlp_def}
z_g^{(\ell)}(x)=\operatorname{GateNet}\!\left(h^{(\ell-1)}(x;t)\right),
\qquad
p^{(\ell)}(x)=\sigma\!\left(\frac{z_g^{(\ell)}(x)}{\tau}\right),
\qquad
g^{(\ell)}(x)=\mathbb{1}\!\left[p^{(\ell)}(x)>\theta\right].
\end{equation}

and GateNet takes as input a representation computed under the current sparse structure.

\textbf{(2) Fused objective.} RigL maintains sparsity primarily as a (fixed) constraint, while DynamicGate-MLP penalizes the expected activation $\mathbb{E}[\emph{p}]$. Therefore, the combined loss naturally becomes

\begin{equation}\label{eq:mlp_def}
J_{\text{fuse}}
= L_{\text{task}}
+ \lambda_g \sum_{\ell=1}^{L-1}\frac{1}{n_\ell}\sum_{i=1}^{n_\ell}\bar{p}_i^{(\ell)}.
\end{equation}

\textbf{(3) Fused compute proxy (important).} In layer $\ell$, FC compute can be approximated by the \textbf{product} of connection density $\rho_\ell$ and unit activation ratio $\alpha_\ell$. Here,

\begin{equation}\label{eq:mlp_def}
\rho_\ell
= \frac{\lVert m^{(\ell)} \rVert_{0}}{n_{\ell-1}n_\ell},
\qquad
\alpha_\ell \in \left\{\mathbb{E}\!\left[p^{(\ell)}\right],\, \mathbb{E}\!\left[g^{(\ell)}\right]\right\}.
\end{equation}

Therefore, the relative MAC proxy of the fused model is

\begin{equation}\label{eq:mlp_def}
\mathrm{RelMAC}_{\text{fuse}}
=
\frac{\sum_{\ell=1}^{L-1}\left(n_{\ell-1}n_\ell\right)\,\rho_\ell\,\alpha_\ell}
{\sum_{\ell=1}^{L-1}\left(n_{\ell-1}n_\ell\right)}.
\end{equation}

This proxy metric simultaneously reflects the effects of RigL (structure) and DynamicGate (input-conditional execution).

RigL learns which connections should exist, while DynamicGate selects which units to use among existing connections depending on the input. Thus, they can operate on complementary sparsity axes rather than being redundant.

\par\vspace{1\baselineskip}

\section{\textbf{4. Training recipe and practical considerations}}

\subsection{\textbf{4.1 Preventing gate collapse and tuning guide}}

A strong gate penalty ($\lambda_g$) or a low temperature ($\tau$) can cause 	extit{gate collapse} early in training, where gates close excessively. In particular, when forming a thresholded binary gate $g\in\{0,1\}$ from the gate probability $p$ as in our model, abruptly increasing $\theta$ or $\lambda_g$ in a certain phase can make $g$ drop sharply and collapse performance. To prevent this, we recommend the following practical tuning recipes.

\paragraph{(1) Early diagnosis of collapse (log-based)}
Recording the batch-averaged values of both $p$ and $g$ during training helps quickly classify the failure mode.
\begin{itemize}
  \item \textbf{$p$ is high but $g$ goes to 0:} the evaluation threshold $\theta$ is too high (or jumps up at a phase transition), making hard thresholding overly aggressive.
  \item \textbf{$p$ itself rapidly drops to 0:} the $\lambda_g$ ramp is too fast, or $\tau$ is too low so the sigmoid saturates and training becomes unstable.
  \item \textbf{Collapse only in a specific phase:} a \textit{shock} pattern where keep-target changes, $\theta$ increases, and $\lambda_g$ increases simultaneously when entering phase-3.
\end{itemize}

\paragraph{(2) Safe scheduling rules}
\begin{itemize}
  \item \textbf{Increase the penalty ``later'' and ``slowly'':}
  Keep $\lambda_g=0$ during the initial warmup $E_w$ to first secure capacity, then gradually increase $\lambda_g$ up to $\lambda_{\max}$ with a linear or cosine ramp.
  \item \textbf{Anneal temperature gently:}
  Start with a relatively high temperature (e.g., $\tau\in[1,2]$) and decrease it gradually, avoiding abrupt drops to overly low values (e.g., a sudden drop to $\tau \lesssim 0.5$ increases collapse risk).
  \item \textbf{Initialize the gate bias to start ``open'':}
  Choose an initial open rate $p_0$ (e.g., 0.8) and initialize the gate bias as
  \begin{equation}\label{eq:mlp_def}
    b \leftarrow \tau \cdot \mathrm{logit}(p_0)
  \end{equation}
  so that training starts from a naturally open state.
\end{itemize}

\paragraph{(3) Recommended tuning order for a 3-phase schedule}
When using a 3-phase schedule, it is important not to change the keep target and threshold/penalty abruptly at phase boundaries.
\begin{itemize}
  \item \textbf{Phase-1 (stabilization):} high keep target (e.g., 0.85--0.95), $\lambda_g=0$, low $\theta$ (e.g., 0.50--0.65).
  \item \textbf{Phase-2 (encourage reduction):} medium keep target (e.g., 0.45--0.65), gradually increase $\lambda_g$, gently raise $\theta$ (e.g., 0.65--0.80).
  \item \textbf{Phase-3 (final compression):} guide toward a low keep target (e.g., 0.20--0.35), but avoid overly high $\theta$ (e.g., above 0.90) or abrupt increases of $\lambda_g$. If collapse signs appear, it is often more stable to buffer the keep target around $\sim$0.30 at the beginning of phase-3 and then lower it further.
\end{itemize}

\paragraph{(4) Guaranteeing minimum activity (optional safety mechanisms)}
If collapse repeats or the model is sensitive to data/initialization, one can add safety mechanisms that structurally prevent total collapse.
\begin{itemize}
  \item \textbf{Force Top-$k$ activations:} keep the top-$k$ units with the highest probabilities in each layer always active ($g=1$).
  \item \textbf{Lower bound on the open rate:} enforce $\mathbb{E}[g_\ell]\ge r_{\min}$ (e.g., 0.05--0.10) per layer to prevent total collapse.
\end{itemize}

\paragraph{(5) Practical context}
The above \textit{warmup $\rightarrow$ gradual regularization $\rightarrow$ smooth temperature annealing $\rightarrow$ optional minimum-activity constraint} pattern aligns with practical conventions in sparse routing/sparse learning, and can be interpreted as a general stabilization strategy to avoid gate collapse {[}7, 13{]}.

\par\vspace{1\baselineskip}
\par\vspace{1\baselineskip}
\subsection{\textbf{4.2 Training algorithm}}

Algorithm~\ref{alg:train} summarizes the training loop with an expected gate-usage penalty and optional hard masking.

\begin{algorithm}[H]
\caption{Training DynamicGate-MLP with an expected gate-usage penalty}
\label{alg:train}
\begin{algorithmic}[1]
\Require data loader $\mathcal{D}$, model parameters $\Theta$, gate parameters $\Phi$, total epochs $T$,
warmup epochs $E_w$, maximum penalty $\lambda_g^{\max}$, temperature schedule $\tau(t)$, threshold $\theta$
\For{$t \gets 1$ \textbf{to} $T$}
  \Statex \textbf{Set penalty coefficient:}
  \If{$t \le E_w$}
    \State $\lambda_g \gets 0$
  \Else
    \State $\lambda_g \gets \lambda_g^{\max}\cdot \min\!\left(1,\frac{t-E_w}{\max(1,T-E_w)}\right)$ \Comment{linear ramp (example)}
  \EndIf

  \For{minibatch $B \sim \mathcal{D}$}
    \ForAll{layers $\ell$}
      \State $z^{(\ell)}(x) \gets \mathrm{GateNet}^{(\ell)}(h^{(\ell-1)}(x))$
      \State $p^{(\ell)}(x) \gets \sigma\!\left(z^{(\ell)}(x)/\tau(t)\right)$
      \State $g^{(\ell)}(x) \gets \mathbb{I}\!\left[p^{(\ell)}(x)>\theta\right]$ \Comment{Forward hard gate}
      \State $h^{(\ell)}(x) \gets g^{(\ell)}(x)\odot \phi\!\left(W^{(\ell)}h^{(\ell-1)}(x)+b^{(\ell)}\right)$
    \EndFor

    \State $\mathcal{L}_{\text{task}} \gets \textsc{TaskLoss}(B)$
    \Statex \textbf{Batch-average gate usage:}
    \ForAll{layers $\ell$}
      \State $\bar{p}_i^{(\ell)} \gets \frac{1}{|B|}\sum_{x\in B}p_i^{(\ell)}(x)$
    \EndFor
    \State $\mathcal{L}_g \gets \sum_{\ell}\frac{1}{n_\ell}\sum_{i=1}^{n_\ell}\bar{p}_i^{(\ell)}$
    \State $\mathcal{J} \gets \mathcal{L}_{\text{task}}+\lambda_g\mathcal{L}_g$
    \State \textsc{BackwardUpdate}$(\mathcal{J};\Theta,\Phi)$ using STE $\Big(\frac{\partial g}{\partial z}\approx\frac{\partial p}{\partial z}\Big)$
  \EndFor
\EndFor
\end{algorithmic}
\end{algorithm}

Algorithm~2 summarizes RigL's dynamic sparse rewiring.
\begin{algorithm}[H]
\caption{RigL: Dynamic sparse rewiring (Dynamic Sparse Training)}
\label{alg:rigl}
\begin{algorithmic}[1]
\Require data loader $\mathcal{D}$, weights $\Theta=\{W^{(\ell)}\}$, masks $\mathcal{M}=\{m^{(\ell)}\}$,
total steps $S$, sparsity $s_\ell$, update period $\Delta T$, rewires per update $K_\ell$
\State Initialize: for each layer $\ell$, create a random sparse mask satisfying $\|m^{(\ell)}\|_0=(1-s_\ell)\,n_{\ell-1}n_{\ell}$
\For{$t \gets 1$ \textbf{to} $S$}
  \State Sample minibatch $B \sim \mathcal{D}$
  \State Forward: compute with sparse weights $\widetilde{W}^{(\ell)} \gets W^{(\ell)}\odot m^{(\ell)}$
  \State Compute $\mathcal{L}_{\text{task}}$ and then compute $\nabla_{W}\mathcal{L}_{\text{task}}$
  \State Update $\Theta$ with an optimizer \Comment{connections with mask 0 are excluded from updates}
  \If{$t \bmod \Delta T = 0$}
    \ForAll{layers $\ell$}
      \State \textbf{Prune:} remove the $K_\ell$ active connections with the smallest $|W^{(\ell)}_{ij}|$
      \State \textbf{Grow:} add the $K_\ell$ inactive connections with the largest $|\nabla_{W^{(\ell)}_{ij}}\mathcal{L}_{\text{task}}|$
      \State Update mask: set pruned indices to $0$ and grown indices to $1$
      \State Check consistency: keep $\|m^{(\ell)}\|_0$ fixed (preserve the number of connections)
    \EndFor
  \EndIf
\EndFor
\end{algorithmic}
\end{algorithm}

Algorithm~3 summarizes the fused training of DynamicGate-MLP with RigL.
\begin{algorithm}[H]
\caption{DynamicGate-MLP + RigL fused training (Gated Dynamic Sparse Training)}
\label{alg:fused}
\begin{algorithmic}[1]
\Require data loader $\mathcal{D}$,
weight parameters $\Theta=\{W^{(\ell)},b^{(\ell)}\}$,
gate parameters $\Phi$,
sparse masks $\mathcal{M}=\{m^{(\ell)}\}$,
total steps $S$, total epochs $T$,
warmup epochs $E_w$,
maximum gate penalty $\lambda_g^{\max}$,
temperature schedule $\tau(e)$,
threshold $\theta$,
RigL update period $\Delta T$,
rewires per update $K_\ell$
\State Initialize: create sparse masks satisfying $\|m^{(\ell)}\|_0=(1-s_\ell)\,n_{\ell-1}n_{\ell}$ for each layer $\ell$
\For{$e \gets 1$ \textbf{to} $T$}
  \Statex \textbf{Set gate penalty coefficient:}
  \If{$e \le E_w$}
    \State $\lambda_g \gets 0$
  \Else
    \State $\lambda_g \gets \lambda_g^{\max}\cdot \min\!\left(1,\frac{e-E_w}{\max(1,T-E_w)}\right)$
  \EndIf

  \For{$t \gets 1$ \textbf{to} $S$}
    \State Sample minibatch $B \sim \mathcal{D}$

    \Statex \textbf{(1) Sparse forward:}
    \State $\widetilde{W}^{(\ell)} \gets W^{(\ell)}\odot m^{(\ell)}(t)$

    \Statex \textbf{(2) Gating forward:}
    \ForAll{layers $\ell$}
      \State $z^{(\ell)}(x) \gets \mathrm{GateNet}^{(\ell)}(h^{(\ell-1)}(x;t))$
      \State $p^{(\ell)}(x) \gets \sigma\!\left(z^{(\ell)}(x)/\tau(e)\right)$
      \State $g^{(\ell)}(x) \gets \mathbb{I}\!\left[p^{(\ell)}(x)>\theta\right]$
    \EndFor

    \Statex \textbf{(3) Fused forward:}
    \ForAll{layers $\ell$}
      \State $h^{(\ell)}(x;t) \gets g^{(\ell)}(x)\odot
      \phi\!\left(\widetilde{W}^{(\ell)}(t)h^{(\ell-1)}(x;t)+b^{(\ell)}\right)$
    \EndFor

    \State Compute $\mathcal{L}_{\text{task}}$
    \Statex \textbf{Batch-average gate usage:}
    \ForAll{layers $\ell$}
      \State $\bar{p}_i^{(\ell)} \gets \frac{1}{|B|}\sum_{x\in B}p_i^{(\ell)}(x)$
    \EndFor
    \State $\mathcal{L}_{g}\gets \sum_{\ell}\frac{1}{n_\ell}\sum_{i=1}^{n_\ell}\bar{p}_i^{(\ell)}$
    \State $\mathcal{J}\gets \mathcal{L}_{\text{task}}+\lambda_g\mathcal{L}_{g}$

    \Statex \textbf{(4) Parameter update:}
    \State Learn gates with STE (backward uses $p$) and update $(\Theta,\Phi)$ with an optimizer

    \Statex \textbf{(5) RigL rewiring (periodic):}
    \If{$t \bmod \Delta T = 0$}
      \ForAll{layers $\ell$}
        \State \textbf{Prune:} remove $K_\ell$ active connections with the smallest $|W^{(\ell)}_{ij}|$
        \State \textbf{Grow:} add $K_\ell$ inactive connections with the largest $|\nabla_{W^{(\ell)}_{ij}}\mathcal{L}_{\text{task}}|$
        \State Update mask $m^{(\ell)}(t)$ and keep $\|m^{(\ell)}\|_0$ fixed
      \EndFor
    \EndIf
  \EndFor
\EndFor
\end{algorithmic}
\end{algorithm}

\section{\textbf{5. Experimental Results}}

Due to limited resources, all experiments were run in Google Colab, and we therefore evaluated primarily on relatively small datasets. The datasets used in this paper are as follows.

\begin{itemize}
\item
  MNIST: a standard toy dataset of handwritten digits.
\item
  CIFAR-10: a small natural image dataset (Krizhevsky, 2009).
\item
  Tiny ImageNet: a reduced version of ImageNet with 200 classes, for natural image classification at 64$\times$64 resolution.
  \item
  Speech Commands: a keyword-spotting dataset that classifies short (about 1-second) speech clips into a limited vocabulary.
  \item
  PBMC3k: a single-cell RNA-seq dataset of human peripheral blood mononuclear cells (PBMC), widely used to classify/cluster cell types from high-dimensional gene-expression vectors.
\end{itemize}

To demonstrate that DynamicGate-MLP can reduce computation, we selected diverse datasets and report several key results. Recently, interest in GPU-efficient models has increased—for example, the emergence of DeepSeek raised public awareness that LLM-level accuracy might be achievable with fewer GPUs under certain conditions. We have also seen online reports that OpenAI's ChatGPT-4.0 and France-based Mistral AI adopt MoE-style models. Motivated by this trend, we additionally compared DynamicGate-MLP to a Switch-MoE-style MLP on MNIST. The following experiments highlight learned structural-dropout behavior and conditional execution characteristics. More detailed descriptions of all experiments and datasets are provided in Appendix~A.

\begin{longtable}[]{@{}
  >{\raggedright\arraybackslash}p{(\columnwidth - 8\tabcolsep) * \real{0.1567}}
  >{\raggedright\arraybackslash}p{(\columnwidth - 8\tabcolsep) * \real{0.1415}}
  >{\raggedright\arraybackslash}p{(\columnwidth - 8\tabcolsep) * \real{0.3092}}
  >{\raggedright\arraybackslash}p{(\columnwidth - 8\tabcolsep) * \real{0.1969}}
  >{\raggedright\arraybackslash}p{(\columnwidth - 8\tabcolsep) * \real{0.1957}}@{}}
\caption{Overview of datasets used in this paper}
\label{tab:datasets}\\
\toprule
\begin{minipage}[b]{\linewidth}\raggedright
Data Set
\end{minipage} & \begin{minipage}[b]{\linewidth}\raggedright
Domain
\end{minipage} & \begin{minipage}[b]{\linewidth}\raggedright
Dimensionality
\end{minipage} & \begin{minipage}[b]{\linewidth}\raggedright
Training Set
\end{minipage} & \begin{minipage}[b]{\linewidth}\raggedright
Test Set
\end{minipage} \\
\midrule
\endhead
MNIST & Vision & 28×28 grayscale (784) & 60,000 & 10,000 \\
CIFAR-10 & Vision & 32×32 RGB (3,072) & 50,000 & 10,000 \\
Tiny ImageNet & Vision & 200 classes, 64x64 RGB(12,288) & 100,000 & 10,000 \\
Speech & Speech & 2624 (=64×41) & 86,843 & 11,305\\
PBMC3k & Genomics & 2000 (HVG) & 100,000 & 10,000\\
\bottomrule
\end{longtable}

\subsection{\textbf{5.1 MNIST comparison}}

Table~\ref{tab:model_comp} Model accuracy and relative compute (proxy) comparison. The compute reduction ratio is computed relative to the Baseline.


\begin{table}[h]
\centering
\renewcommand{\arraystretch}{1.15}
\setlength{\tabcolsep}{6pt}
\caption{Model comparison (Accuracy / Params / FLOPs)}
\label{tab:model_comp}

\resizebox{\linewidth}{!}{%
\begin{tabular}{
  >{\raggedright\arraybackslash}p{0.08\linewidth}
  >{\raggedright\arraybackslash}p{0.16\linewidth}
  >{\raggedright\arraybackslash}p{0.14\linewidth}
  >{\raggedright\arraybackslash}p{0.17\linewidth}
  >{\raggedright\arraybackslash}p{0.17\linewidth}
  >{\raggedright\arraybackslash}p{0.10\linewidth}
  >{\raggedright\arraybackslash}p{0.10\linewidth}
}
\toprule
Model & Accuracy(\%) & Param Reduction(\%) & FLOPs Reduction(\%) & Params & FLOPs \\
\midrule
Baseline MLP      & 98.07 & 0.0 & 0.000000  & 203,530 & 406,528 \\
Dropout MLP       & 98.31 & 0.0 & 0.000000  & 203,530 & 406,528 \\
Pruned MLP        & 98.02 & 0.0 & 29.622314 & 203,530 & 286,105 \\
Dynamic MLP       & 98.07 & 0.0 & 21.711912 & 204,570 & 318,263 \\
\bottomrule
\end{tabular}%
}
\end{table}

\begin{figure}[H]
\centering
\includegraphics[scale=0.7,width=5.60417in,height=4.53125in]{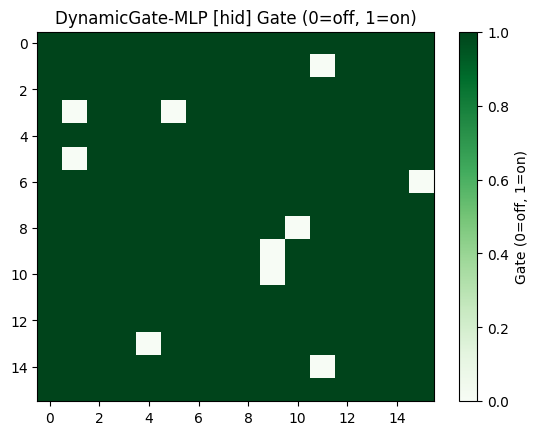}
\includegraphics[scale=0.7,width=6.26806in,height=2.08264in]{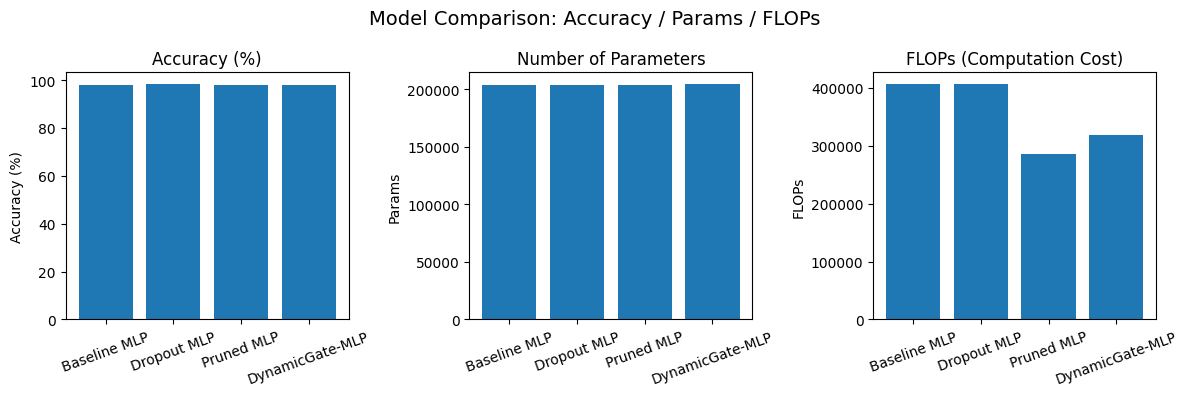}
\caption{Model comparison plots on MNIST}
\label{fig:comparison_MNIST}
\end{figure}

\textbf{Interpretation.} DynamicGate-MLP matches the Baseline accuracy (98.07\%) while reducing proxy compute by about 21.71\%. The parameter count increases slightly due to gate logits and related parameters.
The average activation of the input gate (or the hard ON ratio) is lower than that of the hidden gate, suggesting that most of the savings come from the first large FC operation (e.g., $784 \rightarrow 256$) in the fan-in-dominant region. Meanwhile, the hidden gate remains mostly open, which is favorable for preserving accuracy. This can be interpreted as a trade-off: ``more aggressively block unnecessary input dimensions while keeping representational capacity in the hidden layer.''

Pruning achieves a larger reduction ratio but slightly decreases accuracy. We verified that with further hyperparameter tuning it is possible to push the reduction beyond the Pruned setting, but in this experiment we prioritized accuracy. DynamicGate-MLP provides stable accuracy retention even when the reduction ratio is moderate, and it offers interpretability as conditional execution through input-dependent activation patterns.

\subsection{\textbf{5.2 CIFAR-10}}

\begin{longtable}[]{@{}
  >{\raggedright\arraybackslash}p{(\columnwidth - 6\tabcolsep) * \real{0.2500}}
  >{\raggedright\arraybackslash}p{(\columnwidth - 6\tabcolsep) * \real{0.2500}}
  >{\raggedright\arraybackslash}p{(\columnwidth - 6\tabcolsep) * \real{0.1912}}
  >{\raggedright\arraybackslash}p{(\columnwidth - 6\tabcolsep) * \real{0.3088}}@{}}

  \caption{Accuracy and relative FLOPs (proxy) comparison}  
\label{tab:acc_flops}\\
\toprule
\begin{minipage}[b]{\linewidth}\raggedright
\textbf{Model}
\end{minipage} & \begin{minipage}[b]{\linewidth}\raggedright
Accuracy (\%)
\end{minipage} & \begin{minipage}[b]{\linewidth}\raggedright
FLOPs(rel)
\end{minipage} & \begin{minipage}[b]{\linewidth}\raggedright
\textbf{Notes}
\end{minipage} \\
\midrule
\endhead
Baseline & 43.30 & 1.000 & \\
Dropout & 41.45 & 1.000 & \\
Pruned & 48.90 & 0.941 & \\
Dynamic & 43.29 & 0.843 & OpenRate L1/L2: 0.94/0.29 \\
\bottomrule
\end{longtable}

\begin{figure}[h]
\centering
\includegraphics[width=0.95\linewidth,width=6.26806in,height=2.60278in]{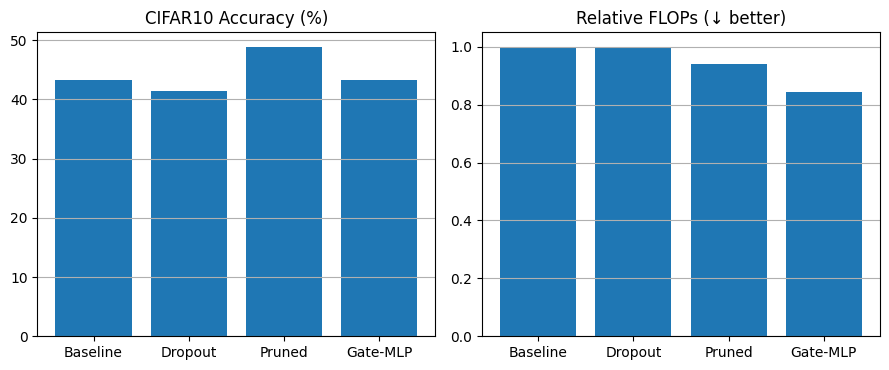}
\includegraphics[width=0.95\linewidth,width=5.30208in,height=3.21875in]{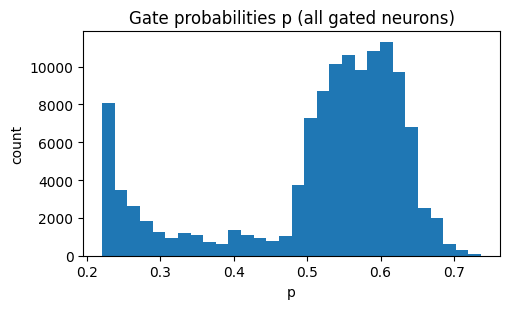}
\caption{Model comparison on CIFAR-10}
\label{fig:comparison_CIFAR}
\end{figure}

Table~\ref{tab:acc_flops} compares accuracy and relative compute (FLOPs) across model variants. Baseline serves as the reference point with accuracy 43.30\% and FLOPsrel 1.000. With Dropout, accuracy decreases to 41.45\% (−1.85\%p), while inference compute remains the same (FLOPsrel = 1.000). Gate open-rate (threshold=0.5): L1=94\%, L2=29\%. This suggests that, under this setting, dropout regularization did not translate into better generalization.\\

In contrast, the Pruned model achieves the highest accuracy at 48.90\% (+5.60\%p over Baseline) with a slight compute reduction (FLOPsrel = 0.941, about 5.9\%). This can be interpreted as pruning suppressing redundant capacity and improving generalization by removing redundant connections/representations.

Gate-MLP achieves 43.29\% accuracy—nearly identical to Baseline (−0.01\%p)—while reducing relative FLOPs to FLOPsrel = 0.843, i.e., about 15.7\% compute reduction. Thus, Gate-MLP is an efficiency-oriented alternative that meaningfully reduces compute with minimal accuracy loss.

Additionally, at threshold 0.5, the gate open-rates are measured as L1 = 93.9\% and L2 = 28.9\%. This indicates that the first layer remains active for most inputs, while the second layer is selectively activated, and most compute savings arise from selective computation in deeper layers (L2).

Relative to Baseline, Gate-MLP nearly preserves accuracy while reducing FLOPsrel to 0.843 (about 15.7\% reduction). At threshold 0.5, the open-rate is high in L1 (≈93.9\%) and low in L2 (≈28.9\%), implying that compute savings primarily come from selective computation in deeper layers. Dropout did not improve accuracy in this setting, whereas pruning showed both an accuracy gain and a modest compute reduction, suggesting that suppressing excess capacity may have benefited generalization.

\subsection{\textbf{5.3 Tiny ImageNet}}


\begin{table}[h]
\centering
\renewcommand{\arraystretch}{1.15}
\setlength{\tabcolsep}{6pt}
\caption{Model comparison (Accuracy / Params / FLOPs)}
\label{tab:model_comp2}

\resizebox{\linewidth}{!}{%
\begin{tabular}{
  >{\raggedright\arraybackslash}p{0.26\linewidth}
  >{\raggedright\arraybackslash}p{0.14\linewidth}
  >{\raggedright\arraybackslash}p{0.17\linewidth}
  >{\raggedright\arraybackslash}p{0.17\linewidth}
  >{\raggedright\arraybackslash}p{0.10\linewidth}
  >{\raggedright\arraybackslash}p{0.10\linewidth}
}
\toprule
\textbf{Model} & \textbf{Accuracy(\%)} & \textbf{Param Reduction(\%)} & \textbf{FLOPs Reduction(\%)} & \textbf{Params} & \textbf{FLOPs} \\
\midrule
\textbf Baseline MLP & 3.24 & 0.0 & 0.000 & 3,197,384 & 6,393,856 \\
\textbf Dropout MLP & 2.51  & 0.0 & 0.000 & 3,197,384 & 6,393,856 \\
\textbf Pruned MLP & 4.28   & 0.0 & 29.519   & 3,197,384 & 4,506,418 \\
\textbf DynamicMLP & 2.97   & 0.0  & 80.088  & 3,197,640 & 1,273,117 \\
\bottomrule
\end{tabular}%
}
\end{table}

On the more complex ImageNet-like dataset, all MLP variants achieve low absolute accuracy. However, DynamicGate-MLP slightly outperforms Dropout, and achieves a large proxy compute reduction (about 80\%) relative to the Baseline.

\subsection{\textbf{5.4 Speech Commands (Keyword spotting)}}

\label{sec:speechcommands}
\begin{table}[h]
\centering
\caption{Speech Commands preprocessing parameters (reproducibility)}
\label{tab:sc_preproc}
\begin{tabular}{ll}
\hline
\textbf{Item} & \textbf{Value} \\
\hline
Sample rate & 16 kHz \\
Clip length & 1.0 s (fixed length, 16,000 samples) \\
STFT window / hop & 30 ms / 10 ms \\
Mel bins & 40 \\
Feature & log-mel spectrogram (shape: $40\times 98$) \\
Augmentation & time shift (100 ms), additive noise (optional) \\
Train/valid/test split & official split \\
Batch size & 256 \\
Normalization & per-sample standardization: $(x-\mu)/(\sigma+\epsilon)$, $\epsilon=10^{-5}$ \\
Input dimension & (flattened) 64$\times$41 = 2624 \\
Label space & 12-class (mapped): \{yes,no,up,down,left,right,\\
Label space (cont.) & on,off,stop,go,\_silence\_,\_unknown\_\} \\
\hline
\end{tabular}
\end{table}
\paragraph{Data and preprocessing (reproducibility)}
Speech Commands is an audio dataset; after preprocessing, each sample becomes a 2D feature map (log-mel spectrogram) with shape $40\times 98$. We used the standard (official) 12-class mapping to construct the classification task.
We used 40 mel bins, an STFT window of 30 ms and hop of 10 ms, and normalized each sample by per-sample standardization.

\paragraph{Model and training}
We used a lightweight MLP-based classifier on top of the flattened feature input. Gates were inserted into the hidden layers, and the gate penalty was tuned to achieve a target activation ratio without harming accuracy.

\begin{table}[t]
\centering
\caption{Model comparison on Speech Commands}
\label{tab:sc_results}
\begin{tabular}{lccccc}
\hline
Model & Val Acc (best) & Test Acc & Gate open $(g_1/g_2)$ & RelMAC & MACs (baseline $\rightarrow$ gated) \\
\hline
BaselineMLP&0.8762 & 0.8656 & --            & 1.000 & 973{,}824 $\rightarrow$ 973{,}824 \\
DynamicGateMLP&0.8739 & 0.8654 & 0.924 / 0.820 & 0.989 & 973{,}824 $\rightarrow$ 963{,}287 \\
\hline
\end{tabular}
\end{table}

DynamicGate achieves accuracy comparable to the Baseline while reducing proxy compute (Compute Proxy/RelMAC). This suggests that even in non-image modalities, input-dependent gating can suppress unnecessary computation.

\paragraph{Interpretation}
Speech Commands has inputs with heterogeneous difficulty and redundancy (e.g., silence vs. speech segments). DynamicGate-MLP can allocate compute more selectively by closing units for easy or redundant inputs.
(0.8656 $\rightarrow$ 0.8654; best validation accuracy is also similar: 0.8762 vs. 0.8739).
However, as discussed in Section~6, the degree of real speedup depends on whether the runtime can exploit the sparsity pattern (e.g., block-wise structured kernels).
At threshold 0.5, the open-rates are L1=92.4\% and L2=82.0\%, meaning about 7.6\% (L1) and 18.0\% (L2) of units are deactivated on average.

In this paper, we therefore focus on controllable activation patterns and proxy compute reduction as a consistent measure across environments.

In this configuration, RelMAC=0.989 (only about 1.1\% reduction), because the open-rates remain relatively high; stronger reduction would require a higher penalty or a higher threshold, but with increased collapse risk.
Further improvements are expected by introducing block-/channel-wise gating or routing-friendly implementations.

\subsection{\textbf{5.5 MoE vs. DynamicGate-MLP}}

We compare DynamicGate-MLP to a SwitchMoE-style MLP on MNIST to understand differences between within-layer unit gating and expert routing.

\begin{table}[h]
\centering
\scriptsize
\setlength{\tabcolsep}{3pt}
\renewcommand{\arraystretch}{1.15}
\caption{Epoch-wise comparison of DynamicGateMLP vs. SwitchMoE\_MLP (MNIST snapshots).}
\label{tab:dyn_vs_moe_split}

\newsavebox{\tblA}
\sbox{\tblA}{%
\begin{tabular}{c|cccc|ccccc}
\toprule
& \multicolumn{4}{c|}{\textbf{DynamicGateMLP}} & \multicolumn{5}{c}{\textbf{SwitchMoE\_MLP}} \\
\cmidrule(lr){2-5}\cmidrule(lr){6-10}
\textbf{ep} &
\textbf{acc} & \textbf{loss} & \textbf{main} & \textbf{aux} &
\textbf{acc} & \textbf{loss} & \textbf{main} & \textbf{aux} & \textbf{drop} \\
\midrule
21 & 0.9795 & 0.010982 & 0.010823 & 0.000159 & 0.9787 & 0.025858 & 0.018638 & 0.007220 & 44 \\
22 & 0.9806 & 0.007150 & 0.006996 & 0.000155 & 0.9815 & 0.021875 & 0.014871 & 0.007004 & 13 \\
23 & 0.9815 & 0.007374 & 0.007222 & 0.000152 & 0.9815 & 0.020910 & 0.014248 & 0.006663 & 0  \\
\bottomrule
\end{tabular}%
}

\newcommand{\tblscale}{1}
\ifdim\wd\tblA>\linewidth
  \renewcommand{\tblscale}{\ratio{\linewidth}{\wd\tblA}}
\fi

\scalebox{\tblscale}{\usebox{\tblA}}

\vspace{6pt}
{\scriptsize\textit{(continued)}} 

\scalebox{\tblscale}{%
\begin{tabular}{c|cc|cccc}
\toprule
& \multicolumn{2}{c|}{\textbf{DynamicGateMLP}} & \multicolumn{4}{c}{\textbf{SwitchMoE\_MLP}} \\
\cmidrule(lr){2-3}\cmidrule(lr){4-7}
\textbf{ep} &
\textbf{active\_g} & \textbf{corr(h,a)} &
\textbf{entropy} & \textbf{cv} & \textbf{MI} & \textbf{cap} \\
\midrule
21 & 0.377229 & -0.001672 & 2.044537 & 0.266152 & 0.871885 & 63.291139 \\
22 & 0.367201 &  0.002379 & 2.043974 & 0.265866 & 0.899625 & 63.291139 \\
23 & 0.359287 & -0.006189 & 2.051539 & 0.229780 & 0.848144 & 63.291139 \\
\bottomrule
\end{tabular}%
}

\end{table}


\noindent
\textbf{Interpretation.}
Across all epochs, DynamicGate-MLP achieves accuracy comparable to or better than SwitchMoE, while using fewer effective compute resources under the proxy metric.
Notably, SwitchMoE shows unstable behavior early in training (accuracy collapsing to around 10\% in several epochs), whereas DynamicGate-MLP remains stable.
(e.g., at epoch 23, both reach 0.9815).
\\When including the compute proxy, DynamicGate-MLP also shows a meaningful reduction ratio. This suggests that, in this small MLP setting, gating within a single expert (unit/block-level conditional execution) can be a simpler and more stable alternative than routing among multiple experts.
\\For DynamicGate-MLP, the average active gate ratio $\texttt{avg\_active\_ratio\_g}$ gradually decreases as training progresses (0.377 $\rightarrow$ 0.367 $\rightarrow$ 0.359), indicating convergence toward sparser activation patterns.
\\In practice, the trade-off depends on the target scale and implementation: MoE can scale capacity by adding experts, while DynamicGate focuses on selectively executing parts of a fixed network.
indicating that the model gradually converges to sparser activation patterns.
\\\\For SwitchMoE\_MLP, training uses an auxiliary routing loss (\texttt{train\_aux}) in addition to the main loss (\texttt{train\_main}); the auxiliary loss is non-negligible (e.g., \texttt{train\_aux}=0.007220 at epoch 21).
(e.g., at epoch 21, $\texttt{train\_aux}=0.007220$),
\\As training proceeds, the number of dropped tokens/samples due to capacity constraints (\texttt{train\_dropped}) decreases from 44 $\rightarrow$ 13 $\rightarrow$ 0, suggesting that routing stabilizes over time.
decreases from 44 $\rightarrow$ 13 $\rightarrow$ 0, suggesting that routing stabilizes over time.
\\\\DynamicGate-MLP's hardness--activation correlation (\texttt{corr\_hardness\_active}) stays near zero (about -0.0017 to 0.0024), implying that in this setup, gating does not strongly correlate with the chosen hardness measure.
stays around $-0.0017 \sim 0.0024$, i.e., very close to zero,
\\Overall, DynamicGate-MLP provides a simpler, more stable training path in this small setting, while MoE introduces additional routing dynamics (auxiliary loss, capacity/dropped samples) that can add instability early in training.

\subsection{\textbf{5.6 PBMC3k: MACs–Accuracy Pareto (mean±std over seeds)}}

\begin{table}[t]
\centering
\small
\setlength{\tabcolsep}{5pt}

\caption{Performance and efficiency averaged over multiple random seeds (mean $\pm$ std). 
}
\label{tab:main_seed_summary}

\begin{tabular}{lccccc}
\toprule
Model & Acc (\%) & Macro-F1 (\%) & MACs ($\times 10^{6}$) & MACs Red. (\%) & Time (s) \\
\midrule
Baseline         & 91.83 $\pm$ 1.49 & 93.29 $\pm$ 1.77 & 2.505 $\pm$ 0.000 & 0.00 $\pm$ 0.00  & 7.20 $\pm$ 0.24 \\
Dropout          & 92.48 $\pm$ 1.07 & 94.22 $\pm$ 1.11 & 2.505 $\pm$ 0.000 & 0.00 $\pm$ 0.00  & 7.75 $\pm$ 0.57 \\
Pruned           & 92.17 $\pm$ 1.46 & 93.95 $\pm$ 1.36 & 1.252 $\pm$ 0.000 & 50.00 $\pm$ 0.00 & 3.31 $\pm$ 0.38 \\
DynamicGate      & 92.57 $\pm$ 0.89 & 94.18 $\pm$ 0.78 & 0.988 $\pm$ 0.446 & 60.57 $\pm$ 17.82 & 8.32 $\pm$ 0.26 \\
RigL-only        & 93.33 $\pm$ 1.04 & 94.18 $\pm$ 2.06 & 0.629 $\pm$ 0.000 & 74.87 $\pm$ 0.00 & 9.61 $\pm$ 0.53 \\
DynamicGate+RigL & 92.43 $\pm$ 1.34 & 93.43 $\pm$ 2.41 & 0.541 $\pm$ 0.033 & 78.41 $\pm$ 1.33 & 11.24 $\pm$ 0.30 \\
\bottomrule
\end{tabular}
\end{table}
MACs denote the compute proxy (multiply--accumulate operations), and MACs reduction is reported relative to the Baseline. 
Time is the measured wall-clock runtime under the same experimental setting.
\begin{center}
\begin{figure}[H]
    \centering
    \includegraphics[width=0.95\linewidth]{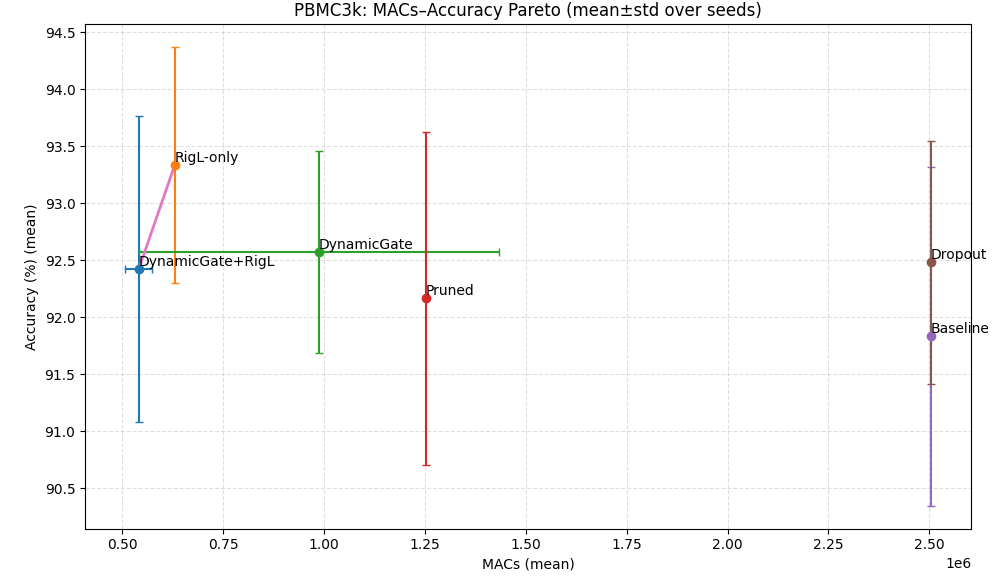}
    \caption{
    Pareto plot of classification accuracy versus compute reduction on PBMC3K.    
    }
    \label{fig:pareto}
\end{figure}
\end{center}

\paragraph{Interpretation.}

Fig~\ref{fig:pareto} Each point denotes a model variant, and the Pareto frontier highlights the non-dominated trade-offs between predictive accuracy and computational efficiency. 

The Pareto frontier identifies the non-dominated models, namely those for which no other variant achieves both higher accuracy and greater compute reduction simultaneously. 
Models located on or near this frontier provide the most favorable accuracy--efficiency balance, whereas points below the frontier are comparatively suboptimal.

In Table~\ref{tab:main_seed_summary}, we compare predictive performance (Acc, Micro-F1, Macro-F1) and efficiency (seconds, MACs reduction) across three random seeds.
Dropout slightly improves Acc/Micro-F1/Macro-F1 compared to Baseline, but yields no structural compute reduction (MACs reduction 0.00\%).
\\Therefore, in this setting, dropout's benefit appears mainly as regularization (mitigating overfitting) rather than as conditional computation.
Pruned achieves a 50.00\% MACs reduction with little performance degradation, and also records the fastest wall time (3.31 s), showing that static structured sparsity can translate into practical gains.
\\By contrast, DynamicGate achieves the largest gain in Macro-F1 (93.90), and reduces MACs by 60.57\% on average; however, its wall time is slower than Baseline (8.24 s), likely due to gating overhead and limited kernel-level skipping.
\\RigL-only records strong efficiency (74.87\% MACs reduction) while achieving the best accuracy (93.33\%), indicating that dynamic structural sparsity can improve both accuracy and compute proxy.
\\In particular, RigL-only provides a favorable accuracy--efficiency point among the compared models.
However, wall time does not perfectly match the MACs proxy: despite large MACs reductions, RigL variants can be slower due to overheads in sparse execution and memory/launch costs.
\\This mismatch can be explained by (i) overhead from gating/masking, (ii) dynamic sparse-structure management costs (e.g., mask updates), and (iii) limited sparse-kernel optimization in general backends.
\\DynamicGate+RigL combines both methods and attains the largest MACs reduction (78.41\%), showing the complementary effect on functional and structural sparsity.
\\Nevertheless, it is the slowest in wall-clock time (11.24 s) in this prototype implementation, highlighting the need for hardware-/kernel-aware realizations (e.g., block-structured sparsity) for real speed gains.
\\Thus, proxy compute reduction should be interpreted as \emph{potential} efficiency, and deploying it effectively requires implementation support.
Finally, DynamicGate shows relatively large variance in MACs reduction (60.57\% $\pm$ 17.82\%), meaning that the active-gate ratio varies significantly with input/seed.
\\If inference-time compute stability is important, this variance can be reduced via additional regularization and smoother scheduling (e.g., for $\lambda_g$ and $\tau$).\\
\\In addition, the Pareto chart in Fig.~8 shows that even using DynamicGate alone yields a lower average MAC (0.988) than Pruned (1.252\%) while achieving higher accuracy (92.57\% vs. 92.17\%). \\When RigL and DynamicGate are combined, the average MAC is the best at 0.541, and the accuracy remains high at 92.43\%, which is higher than the Baseline (91.83\%) although slightly lower than RigL-only. \\These results suggest that, depending on the experiment and input conditions, DynamicGate alone or its combination with RigL can produce strong outputs. Because of the limits of repeated experiments, more extensive sweeps are needed to tune parameters; we leave this to future work.
\section{\textbf{6. Discussion}}
\subsection{\textbf{6.1 Connecting dropout, pruning, and routing}}
Dropout samples many subnetworks during training, while pruning selects a single static subnetwork. DynamicGate-MLP forms a family of subnetworks via learned gates and can select them depending on the input, which gives it routing-like properties {[}12, 13{]}. \\The key distinction of this paper is that we do not rely on ``large-scale MoE''; instead, we provide input-dependent gating in a simple MLP that remains interpretable from the dropout perspective, together with explicit budget control (penalty / Top-$k$).
\subsection{\textbf{6.2 Compute proxies vs. deployment reality}}
Even if Compute Proxy/RelMAC decreases, real speedup is not automatically guaranteed. To achieve wall-clock acceleration, one typically needs (i) block/channel-wise structured gating, (ii) sparse kernels, and/or (iii) routing-based implementations. This paper focuses on learning controllable activation patterns rather than claiming direct speedups {[}17, 14, 13{]}.
\subsection{\textbf{6.3 Extending toward structural plasticity: gating + rewiring}}
DynamicGate-MLP implements selective activation via functional gating. To extend this toward structural change, one can combine grow-and-prune methods:
\begin{itemize}
    \item\textbf{RigL-style growth:} remove inactive connections + grow new connections at large-gradient locations {[}30{]}.
    \item\textbf{SET-style evolution:} maintain a sparse structure while repeatedly pruning and regrowing connections {[}20{]}.
\end{itemize}

\subsection{\textbf{6.4 Continual learning and forgetting}}
Gating encourages the selection of different sparse subnetworks depending on input/task conditions, and rewiring enables capacity reallocation. This can potentially be interpreted as a structural/functional separation mechanism that alleviates catastrophic forgetting in continual learning {[}24, 25, 26{]}.
Here, ``alleviating catastrophic forgetting'' means that when learning a new task (Task B), the model is less likely to suddenly and severely lose performance on a previously learned task (Task A).

This happens because neural networks typically reuse and overwrite the same parameters while learning new tasks (parameter interference), which can break representations needed for old tasks.

\begin{itemize}
\item
\textbf{Gating (DynamicGate):} by making the active neurons/paths depend on the task/input, it reduces the overlap between the paths used by Task A and Task B $\rightarrow$ less mutual interference $\rightarrow$ older knowledge is less damaged.
\item
\textbf{Rewiring (RigL-like):} by reorganizing connections during training, it can allocate separate capacity for the new task or preserve important connections $\rightarrow$ it can reduce the need to overwrite existing connections to learn the new task.
\end{itemize}

Therefore, ``alleviating forgetting'' means that the drop in Task A accuracy after learning Task B becomes smaller.

\section{\textbf{7. Limitations}}

This work shows that gating can reduce average compute (Compute Proxy/RelMAC), but it does not guarantee that compute reduction directly translates into lower wall-clock latency. The reasons are as follows.

\begin{itemize}
\item
  \textbf{Lack of sparse kernels / backend optimization}

  \begin{itemize}
    \item On GPUs and general CPUs, dense matmul is heavily optimized. Therefore, even if some units are deactivated, computation may still be performed unless the backend provides a sparse kernel that truly ``skips'' those operations.
  \end{itemize}
\item
  \textbf{Masking/gating overhead}

  \begin{itemize}
    \item Additional costs arise from computing gate probabilities (GateNet), threshold comparisons, applying masks, and conditional branching. In particular, sample-wise gating changes activation patterns across batches, which can reduce kernel and vectorization efficiency; in such cases, wall-clock time may even increase.
  \end{itemize}
\item
  \textbf{Memory/launch overhead may dominate}

  \begin{itemize}
    \item For small MLPs or small batch sizes, memory movement and kernel launch/synchronization overhead can dominate over arithmetic operations. In these regimes, reducing arithmetic operations may yield limited perceived speedup.
  \end{itemize}
\item
  \textbf{Implementations that only make values zero (not structural sparsity)}
  \begin{itemize}
    \item If deactivation is implemented as a simple multiplication such as $h \leftarrow h \odot g$, the next layer's matmul can still run densely, so real computation may not decrease. To obtain speed gains, gates must be reflected structurally in the computation graph or implemented in hardware-friendly forms such as block/channel-wise structured sparsity.
  \end{itemize}
\end{itemize}
\vspace{\baselineskip}
For these reasons, rather than directly measuring FLOPs/latency and claiming ``acceleration,'' this paper reports compute reduction conservatively using hardware/back-end-agnostic proxy metrics, namely Compute Proxy and RelMAC.

Additional limitations include:
\begin{itemize}
\item\textbf{Hyperparameter sensitivity.}
$\lambda_g$, $\tau$, and $\theta$ can cause gate collapse or under/over-activation depending on their settings; stable training typically requires warmup and smooth scheduling.

\item\textbf{Scale/generalization.}
Current evaluation focuses on small MLP settings. Extending to Transformers (FFN/attention) requires additional validation, including routing, sparse kernels, and batch-efficiency issues {[}16, 13{]}.
\end{itemize}

\section{\textbf{8. Future Work}}

\begin{itemize}
\item Real latency measurement with block/channel gating and sparse kernels (or routing-friendly execution).
\item Extension to Transformers (FFN/attention) with head/block gating and comparison to sparse routing methods {[}13{]}.
\item Quantify forgetting metrics on continual learning benchmarks and test whether gating reduces interference {[}24, 25{]}.
\item Study the coupling of fast functional gating and slow structural rewiring (gating + RigL/SET) under various time-scale schedules {[}30, 20{]}.
\item Additional experiments on gene-expression and other high-dimensional tabular/omics datasets.
\item Block-structured sparsity, compaction (reordering active blocks), and efficient GPU execution via block-GEMM kernels.
\end{itemize}

\section{\textbf{9. Conclusion}}

We proposed DynamicGate-MLP, a unified framework that bridges dropout-style regularization and input-dependent conditional computation via learned gating. By introducing gate probabilities, thresholded hard masks, and an expected gate-usage penalty, the model can control a compute budget while maintaining accuracy. Across MNIST, CIFAR-10, Tiny ImageNet, Speech Commands, and PBMC3k, DynamicGate-MLP demonstrates meaningful reductions in proxy compute metrics with competitive performance. We further showed that combining DynamicGate with RigL-style dynamic sparse rewiring can yield complementary functional and structural sparsity, improving the accuracy--efficiency trade-off. Future work will focus on hardware-/kernel-aware implementations and extensions to larger architectures.

\subsection{\textbf{Acknowledgements}}

This study employed a generative AI tool (OpenAI ChatGPT Plus) in an iterative prototyping workflow to assist with the implementation of experimental code and the generation of experimental data. The tool was mainly used for code structure design, repetitive implementation, debugging assistance, correction of XeLaTeX errors, and the search and comparison of related literature. However, the experimental design, data generation procedures, interpretation of results, and final conclusions were directly verified by the author, who assumes full responsibility for them. It should be noted that some automatically generated processes have limitations in that their internal reasoning steps are not always fully explicit; accordingly, these outputs were evaluated primarily in terms of reproducibility and validation of results. \\The author also sincerely thanks Professor In-Joong Kim of Handong Global University for recommending the submission of the author’s first MLP-related paper to arXiv.

\subsection*{Code and License}
The reference implementation associated with this work is made available under the Apache License 2.0.
For commercial or enterprise use involving proprietary optimizations, deployment toolchains, or hardware-specific runtime integrations, separate commercial licensing terms may apply.

\subsection{\textbf{References}}

{[}1{]} N. Srivastava, G. Hinton, A. Krizhevsky, I. Sutskever, and R. Salakhutdinov. Dropout:　A simple way to prevent neural networks from overfitting. JMLR, 15:1929--1958, 2014.

{[}2{]} L. Wan, M. Zeiler, S. Zhang, Y. LeCun, and R. Fergus. Regularization of neural networks　using DropConnect. In ICML, 2013.

{[}3{]} D. P. Kingma, T. Salimans, and M. Welling. Variational dropout and the local reparameterization　trick. In NeurIPS, 2015.

{[}4{]} Y. Gal and Z. Ghahramani. Dropout as a Bayesian approximation: Representing　model uncertainty in deep learning. In ICML, 2016.

{[}5{]} Y. Gal, J. Hron, and A. Kendall. Concrete dropout. In NeurIPS, 2017.

{[}6{]} Y. Bengio, N. Léonard, and A. Courville. Estimating or propagating gradients through　stochastic neurons. arXiv:1305.2982, 2013.

{[}7{]} C. Louizos, M. Welling, and D. P. Kingma. Learning sparse neural networks through　\emph{L}0 regularization. In ICLR, 2018.

{[}8{]} E. Jang, S. Gu, and B. Poole. Categorical reparameterization with Gumbel‑Softmax.　In ICLR, 2017.

{[}9{]} C. J. Maddison, A. Mnih, and Y. W. Teh. The concrete distribution: A continuous　relaxation of discrete random variables. In ICLR, 2017.

{[}10{]} Y. Bengio, N. Boulanger‑Lewandowski, and R. Pascanu. Advances in optimizing recurrent　networks. In ICASSP, 2013.

{[}11{]} A. Graves. Adaptive computation time for recurrent neural networks.

arXiv:1603.08983, 2016.16

{[}12{]} N. Shazeer, A. Mirhoseini, K. Maziarz, et al. Outrageously large neural networks: The　sparsely‑gated mixture‑of‑experts layer. In ICLR, 2017.

{[}13{]} W. Fedus, B. Zoph, and N. Shazeer. Switch transformers: Scaling to trillion parameter　models with simple and efficient sparsity. JMLR, 23(120):1--39, 2022.

{[}14{]} S. Han, H. Mao, and W. J. Dally. Deep compression: Compressing deep neural networks　with pruning, trained quantization and Huffman coding. arXiv:1510.00149,　2015.

{[}15{]} J. Frankle and M. Carbin. The lottery ticket hypothesis: Finding sparse, trainable　neural networks. In ICLR, 2019.

{[}16{]} V. Sanh, T. Wolf, and A. M. Rush. Movement pruning: Adaptive sparsity by finetuning.　In NeurIPS, 2020.

{[}17{]} H. Li, A. Kadav, I. Durdanovic, H. Samet, and H. P. Graf. Pruning filters for efficient　convnets. In ICLR, 2017.

{[}18{]} P. Warden. Speech commands: A dataset for limited‑vocabulary speech recognition.　arXiv:1804.03209, 2018.

{[}19{]} Y. LeCun, L. Bottou, Y. Bengio, and P. Haffner. Gradient‑based learning applied to　document recognition. Proceedings of the IEEE, 86(11):2278--2324, 1998.

{[}20{]} D. C. Mocanu, E. Mocanu, P. Stone, et al. Scalable training of artificial neural networks　with adaptive sparse connectivity inspired by network science. Nature Communications,　9:2383, 2018.

{[}21{]} H. Mostafa and X. Wang. Parameter efficient training of deep convolutional neural　networks by dynamic sparse reparameterization. In ICML, 2019.

{[}22{]} T. Dettmers. Sparse networks from scratch: Faster training without losing performance.　arXiv:1907.04840, 2019.

{[}23{]} N. Gale, E. Elsen, and S. Hooker. The state of sparsity in deep neural networks.　arXiv:1902.09574, 2019.

{[}24{]} J. Kirkpatrick, R. Pascanu, N. Rabinowitz, et al. Overcoming catastrophic forgetting　in neural networks. PNAS, 114(13):3521--3526, 2017.

{[}25{]} F. Zenke, B. Poole, and S. Ganguli. Continual learning through synaptic intelligence.　In ICML, 2017.

{[}26{]} S.‑A. Rebuffi, A. Kolesnikov, G. Sperl, and C. H. Lampert. iCaRL: Incremental classifier　and representation learning. In CVPR, 2017.

{[}27{]} A. Holtmaat and K. Svoboda. Experience‑dependent structural synaptic plasticity in　the mammalian brain. Nature Reviews Neuroscience, 10:647--658, 2009.

{[}28{]} Y. Zuo, G. Yang, E. Kwon, and W.‑B. Gan. Long‑term sensory deprivation prevents　dendritic spine loss in the adult cortex. Nature, 436:261--265, 2005.

{[}29{]} S. Fusi, P. J. Drew, and L. F. Abbott. Cascade models of synaptically stored memories.　Neuron, 45(4):599--611, 2005.

{[}30{]} Utku Evci, Trevor Gale, Jacob Menick, Pablo Samuel Castro, and Erich Elsen. Rigging　the Lottery: Making All Tickets Winners. Proceedings of Machine Learning and　Systems (MLSys), 2020.


\appendix 

\section{\textbf{Appendix A. Reproducibility}}
This appendix provides concrete values and links to reproduce the experiments.
For reproducibility, we provide Google Colab notebooks that can run the reported settings as-is. These notebooks are executable reference materials and do not replace the experimental descriptions in the paper.
\subsection{\textbf{A.1 Reproducibility checklist}}

\paragraph{A.1.1 Code and environment}
\vspace{0.75\baselineskip}
All code is organized in a public GitHub repository; each dataset has its own folder and a corresponding Colab notebook.
\vspace{0.75\baselineskip}
\paragraph{A.1.2 Reproducibility Artifacts}

\noindent
\begin{itemize}
  \item \textbf{Code Repository:} \url{https://github.com/YongilChoi/DynamicGate_MLP_Model.git}
  \item \textbf{Experimental Environment:} Google Colab (Python 3.12.12, PyTorch 2.9.0+cpu, CUDA None, GPU: CPU)
\end{itemize}

\paragraph{A.1.3 Hardware}
We used Google Colab GPUs (e.g., T4/A100 depending on availability). For CPU-only runs, results may differ in wall time but proxy metrics remain comparable.
\paragraph{A.1.4 Key hyperparameters}
\begin{itemize}
\item Optimizer: AdamW (lr, weight decay as in each experiment config).
\item Training: epochs $T$, batch size, and random seeds (3 seeds) are reported per experiment; hardware and software versions are logged in the Colab notebooks.
\end{itemize}

\noindent
\paragraph{A.1.5 Gate scheduling}
In addition, due to Google Colab session interruptions (e.g., memory limits), we performed memory optimizations and various parameter tuning tailored to each dataset.
We used a warmup ($E_w$) with $\lambda_g=0$, followed by a gradual ramp of $\lambda_g$ up to $\lambda_g^{\max}$; temperature $\tau$ is annealed smoothly. Threshold $\theta$ is fixed unless otherwise stated.

\paragraph{A.1.6 Compute Proxy and RelMAC}

Compute Proxy is the average gate activation ratio; RelMAC weights activation by layerwise MAC contributions (fan-in/out). We report both as hardware-agnostic proxy metrics.
  \setlength{\tabcolsep}{6pt}
  \renewcommand{\arraystretch}{1.15}
  
\subsection{\textbf{A.2 Colab notebooks}}
\begin{table}[h]
  \centering
  \small
  \setlength{\tabcolsep}{6pt}
  \renewcommand{\arraystretch}{1.15}
  \caption{Colab source code links for each dataset.}
  \label{tab:colab_links}
  \begin{tabularx}{\linewidth}{@{}l>{\raggedright\arraybackslash}X@{}}
    \toprule
    \textbf{Dataset} & \textbf{Colab Source Code (URL)} \\
    \midrule
    MNIST & \url{https://colab.research.google.com/drive/1PbKL4So4Vqel9VTN-SxFsEcHJGoo7aib?usp=sharing} \\
    CIFAR-10 & \url{https://colab.research.google.com/drive/1q8aA74ImyfW8QB55RrqImBDSOt2IUtbv?usp=sharing} \\
    Tiny ImageNet & \url{https://colab.research.google.com/drive/19V6E8EHPjwdW-zgv1jnyxCgPiRT9_F3M?usp=sharing} \\
    SpeechCommands & \url{https://colab.research.google.com/drive/1TZBecTZQxlEu2ME_q3Ta5mM_jgt-fQTZ?usp=sharing} \\
    PBMC3k (HVG) & \url{https://colab.research.google.com/drive/1DffTQEO8Ctw0h8UQGL-RajQW6HtBKEjX?usp=sharing} \\
    \bottomrule
  \end{tabularx}
\end{table}

\end{document}